\documentclass[10pt]{article}
\usepackage{rev_sbrn}
\usepackage{times,amsfonts,enumerate,amssymb,amsmath,epsfig,subfigure,bm,cite}
\usepackage[latin1]{inputenc}
\usepackage[OT1]{fontenc}
\usepackage[brazil]{babel}

\usepackage[a4paper, 
hmargin={2cm,1cm}, 
vmargin={2cm,2cm},
footskip=5mm]{geometry}

\begin{document}
\title{AVALIAÇÃO DO MÉTODO DIALÉTICO NA QUANTIZAÇÃO DE IMAGENS MULTIESPECTRAIS}

\author{%
{\bf Wellington Pinheiro dos Santos} \\ 
{\normalsize Núcleo de Engenharia Biomédica, Centro de Tecnologia e Geociências, Universidade Federal de Pernambuco}\\ 
{\normalsize wellington.santos@ieee.org} \\
{\bf Francisco Marcos de Assis}\\
{\normalsize Departamento de Engenharia Elétrica, Universidade Federal de Campina Grande} \\ 
{\normalsize fmarcos@dee.ufcg.edu.br} \\
}
\maketitle 

\thispagestyle{cbrna}
\pagestyle{cbrna}

\setcounter{page}{1}

\noindent{{\bf\large Resumo --} A classificação não supervisionada tem um papel muito importante na análise de imagens multiespectrais, dada a sua capacidade para auxiliar a extração de conhecimento \emph{a priori} de imagens. Algoritmos como k-médias e \emph{fuzzy} c-médias tem sido muito utilizados nessa tarefa. A Inteligência Computacional tem-se mostrado como um importante campo para auxiliar na construção de classificadores otimizados, tanto quanto à qualidade do agrupamento de classes, quanto à avaliação da qualidade da quantização vetorial. Diversos trabalhos têm mostrado que a Filosofia, em especial o Método Dialético, tem servido como importante inspiração para a construção de novos métodos computacionais. Este trabalho apresenta uma avaliação de quatro métodos baseados na Dialética: o Classificador Dialético Objetivo e o Método Dialético de Otimização adaptado à construção de uma versão do k-médias otimizada segundo índices de qualidade de agrupamento, cada um desses em uma versão canônica e em outra versão obtida pela aplicação do Princípio da Máxima Entropia. Esses métodos foram comparados aos métodos k-médias, \emph{fuzzy} c-médias e mapa auto-organizado de Kohonen. Os resultados mostraram que os métodos baseados na Dialética são robustos ao ruído e podem atingir resultados de quantização tão bons quanto aqueles obtidos com o mapa de Kohonen, considerado um quantizador ótimo.}
\\

\noindent{{\bf\large Palavras-chave --} Segmentação de imagens, k-médias, dialética, otimização, princípio da máxima entropia, computação evolucionária.}
\\

\noindent{{\bf\large Abstract --} The unsupervised classification has a very important role in the analysis of multispectral images, given its ability to assist the extraction of a priori knowledge of images. Algorithms like k-means and fuzzy c-means has long been used in this task. Computational Intelligence has proven to be an important field to assist in building classifiers optimized according to the quality of the grouping of classes and the evaluation of the quality of vector quantization. Several studies have shown that Philosophy, especially the Dialectical Method, has served as an important inspiration for the construction of new computational methods. This paper presents an evaluation of four methods based on the Dialectics: the Objective Dialectical Classifier and the Dialectical Optimization Method adapted to build a version of k-means with optimal quality indices; each of them is presented in two versions: a canonical version and another version obtained by applying the Principle of Maximum Entropy. These methods were compared to k-means, fuzzy c-means and Kohonen's self-organizing maps. The results showed that the methods based on Dialectics are robust to noise, and quantization can achieve results as good as those obtained with the Kohonen map, considered an optimal quantizer.}
\\

\\

\section{Introdução}

Diversos trabalhos mostram que é possível construir um algoritmo para classificação não supervisionada de imagens baseado no método dialético \cite{santos2009a,santos2009c,santos2009d,santos2009e,santos2009f}. A principal vantagem dos assim chamados classificadores dialéticos objetivos está na possibilidade de se realizar uma classificação sem saber previamente o número real de classes presente na imagem \cite{santos2008a,santos2008b}, característica compartilhada com diversas redes neurais construtivas \cite{haykin2001,santos2007a,santos2007r2,santos2007b,santos2007c,santos2007d,santos2006a} e pelo método de agrupamento x-médias \cite{pelleg-xmeans}. No entanto, também foi visto que é possível usar o método dialético como inspiração para construir métodos de busca e otimização, podendo-se alcançar resultados interessantes quanto à precisão da otimização e ao número de iterações e de avaliações da função dispendidos até se alcançar um determinado valor limiar de otimização \cite{santos2009b,santos2009i,santos2009j}.

Contudo, aproveitando o mesmo princípio da otimização de funções, é possível construir algoritmos de classificação não supervisionada e agrupamento usando métodos evolucionários de otimização, onde os vetores candidatos à solução são construídos com parâmetros do classificador (no caso do mapa de k-médias, esses parâmetros seriam compostos pelas coordenadas dos pesos dos centróides); assim, avaliar uma determinada função objetivo em um determinado ponto equivaleria a realizar uma classificação usando os parâmetros do classificador passados pelo vetor candidato à solução e, após isso, avaliar a qualidade da classificação segundo algum item de avaliação da validade do agrupamento, no caso de se otimizar um método de agrupamento, como o k-médias, por exemplo. Assim, é possível usar técnicas de otimização em substituição ao algoritmo de treinamento do método de agrupamento a otimizar, resultando classificadores com determinadas características otimizadas.

Neste artigo se busca avaliar o uso dos classificadores dialéticos objetivos quanto à qualidade da quantização, usando para tanto índices de fidelidade como medidas indiretas da distorção de quantização, além de índices de validade de agrupamento. Os resultados são comparados com resultados semelhantes obtidos com classificadores baseados no mapa de k-médias, no mapa \emph{fuzzy} c-médias e no mapa auto-organizado de Kohonen.

Também é proposto um novo método de classificação não supervisionada e agrupamento baseado na otimização do mapa de k-médias usando o método dialético de otimização em função de índices de avaliação da validade do agrupamento. Todos os resultados são gerados a partir da classificação de imagens de ressonância magnética sintéticas e comparados entre si, considerando as diversas versões canônica e com entropia maximizada, e com resultados gerados com classificadores baseados no mapa de k-médias padrão, no mapa auto-organizado de Kohonen, e no mapa \emph{fuzzy} c-médias, usando testes estatísticos.

Este artigo está organizado da maneira que segue: na seção \ref{sec:quanti_metodos} são descritos o método dialético de otimização, índices e medidas de validação de agrupamento, o método da máxima entropia, e métodos de otimização do algoritmo de k-médias, além de outras abordagens de segmentação, como o algoritmo \emph{fuzzy} c-médias; nessa mesma seção são apresentadas as imagens utilizadas; na seção \ref{sec:quanti_resultados} são apresentados os resultados obtidos com a abordagem proposta, enquanto na seção \ref{sec:quanti_conclusoes} são tecidas as conclusões finais. 

\section{Materiais e Métodos} \label{sec:quanti_metodos}

O objetivo principal deste trabalho é mostrar que tanto o classificador dialético objetivo (ODC) quanto o método dialético de busca e otimização (ODM) podem ser utilizados para segmentar imagens multiespectrais, sendo o primeiro um método de classificação e segmentação em si, enquanto o último é utilizado para minimizar uma função de custo de um algoritmo de agrupamento padrão aplicado à segmentação de imagens.

A ideia do uso do ODM na classificação é substituir o algoritmo de treinamento do classificador não supervisionado por um processo de otimização usando o ODM para otimizar o resultado de algum índice de avaliação da qualidade do processo de agrupamento (ou \emph{clustering}) e, em seguida, avaliar o resultado utilizando índices de fidelidade que medem a semelhança entre a imagem quantizada obtida da segmentação e a imagem original.

O método de agrupamento otimizado neste trabalho foi o mapa de k-médias, enquanto as imagens adotadas neste estudo de caso consistem em imagens multiespectrais sintéticas de ressonância magnética. Os resultados em seguida são comparados com aqueles obtidos por outros algoritmos de agrupamento, como o k-médias padrão, o \emph{fuzzy} c-médias e o mapa auto-organizado de Kohonen, quanto à distorção de quantização. Para comparação são utilizados testes estatísticos, como o teste de $\chi^2$, para avaliar quão próximos são os resultados de quantização do ponto de vista global, ou seja, considerando diversos resultados para índices de fidelidade diferentes.

\subsection{Algoritmo de Busca e Otimização}

O método dialético objetivo pode ser adaptado para as tarefas de busca e otimização da seguinte maneira \cite{santos2010a,santos2009b,santos2009h}:
\begin{enumerate}
	\item Definem-se a quantidade inicial $m(0)$ de \emph{polos} integrantes do sistema dialético $\Omega(0)$, juntamente com o \emph{número de fases históricas}, $n_P$, e a \emph{duração de cada fase histórica}, $n_H$. Esses são parâmetros do método dialético de otimização. O número inicial de polos corresponde ao conceito de população inicial das abordagens evolucionárias tradicionais. O número inicial de polos deve ser par, de forma que metade do número de polos seja gerada de forma aleatória, dentro do domínio da função objetivo, enquanto a outra metade é obtida pela geração de polos antítese absoluta.	Assim, do ponto de vista da concepção dialética, têm-se um conjunto inicial de polos composto de pares tese-antítese em contradição antagônica (um dos polos é uma tese, enquanto o outro é sua antítese antagônica), gerando maior dinâmica inicial, ou seja, a luta de polos é mais intensa \cite{vazquez2007,konder2004}. Já do ponto de vista da computação evolucionária, essa estratégia acelera tipicamente em 10\% a convergência do algoritmo de otimização, uma vez que os algoritmos de busca e otimização baseados em computação evolucionária são altamente dependentes da população inicial e, portanto, a probabilidade de encontrar os pontos ótimos utilizando menos iterações pode ser aumentada ao se avaliar tanto um candidato à solução quanto o seu oposto \cite{rahnamayan2007}. Contudo, é importante observar que a convergência prematura pode prejudicar a eficácia do algoritmo \cite{rahnamayan2007}. Além do mais, a rápida convergência de um algoritmo de otimização pode não implicar em encontrar o ponto ótimo mais rápido para todos os casos \cite{rahnamayan2007}. Levando em conta essas considerações, os polos iniciais são definidos da forma que segue:
	\begin{equation}
	  w_{i,j}(0)=
	  \begin{cases}
	    {U(r_j,s_j),} & {1\leq i\leq \frac{1}{2} m(0),}\\
	    {\breve{w}_{i',j}(0),} & {1+\frac{1}{2} m(0) \leq i\leq m(0),}
	  \end{cases}
	\end{equation}
	\begin{equation}
	  \breve{w}_{i,j}=s_j-w_{i,j}+r_j,
	\end{equation}
	para $i'=i-\frac{1}{2}m(0)$, $1\leq i\leq m(0)$ e $1\leq j\leq n$, onde $n$ é a dimensionalidade do problema de otimização, $U(r_j,s_j)$ é um número aleatório uniformemente distribuído no intervalo $[r_j,s_j]$ e $S=[r_1,s_1]\times [r_2,s_2]\times \dots \times [r_n,s_n]$, desde que $s_j>r_j$ e $s_j,r_j\in \mathbb{R}$. 
	\item Enquanto não se atinge um máximo de $n_P$ fases históricas e a força hegemônica histórica não é maior do que um dado limiar superior de força (estimativa inicial do valor máximo da função objetivo), $f_H(t)<f_{\sup}$ (critério para se considerar o máximo da função objetivo atingido), repete-se:
	\begin{description}
		\item[Evolução:] Enquanto não se atinge um máximo de $n_H$ iterações e $f_H(t)<f_{\sup}$, os polos são ajustados segundo a seguinte expressão:
		\begin{equation}
		  \mathbf{w}_i(t+1)=\mathbf{w}_i(t) + \Delta \mathbf{w}_{C,i}(t) + \Delta \mathbf{w}_{H,i}(t), 
		\end{equation}
		para
		\begin{equation}
		  \Delta \mathbf{w}_{C,i}(t) = \eta_C(t)(1-\mu_{C,i}(t))^2 (\mathbf{w}_C(t)-\mathbf{w}_i(t)),
		\end{equation}
		\begin{equation}
		  \Delta \mathbf{w}_{H,i}(t) = \eta_H(t)(1-\mu_{H,i}(t))^2 (\mathbf{w}_H(t)-\mathbf{w}_i(t)),
		\end{equation}
		\begin{equation}
		  0<\eta_C(t)<1,
		\end{equation}
		\begin{equation}
		  0<\eta_H(t)<1,
		\end{equation}
		onde $\eta_C(0)=\eta_H(0)=\eta_0$ e $0<\eta_0<1$. Os termos $\Delta \mathbf{w}_{C,i}(t)$ e $\Delta \mathbf{w}_{H,i}(t)$ modelam as influências das hegemonias presente e histórica, nesta ordem, sobre o $i$-ésimo polo, enquanto $\eta_C(t)$ e $\eta_H(t)$ são os respectivos passos de atualização dos polos, atualizados a cada iteração e a cada fase histórica, respectivamente, de forma que
		\begin{equation}
		  \eta_C(t+1)=\alpha \eta_C(t),
		\end{equation}
		ao final de cada iteração e
		\begin{equation}
		  \eta_H(t+1)=\alpha \eta_H(t),
		\end{equation}
		ao final de cada fase histórica, para $\alpha<1$ (tipicamente, $\alpha=0,9999$). O decrescimento dos passos ao longo do tempo, embora lento, é efetuado para incrementar a busca e, portanto, facilitar a convergência do algoritmo, tal como em algumas redes neurais artificiais, garantindo assim a convergência assintótica do algoritmo \cite{haykin2001}. Os termos $\mu_{C,i}$ e $\mu_{H,i}$ são chamados de \emph{pertinência presente} e \emph{per\-ti\-nên\-cia histórica}, respectivamente, sendo definidos da forma que segue, baseada nas funções de per\-ti\-nên\-cia da versão clássica do classificador \emph{fuzzy} c-médias \cite{hung2006,chen2006,windischberger2003,alexiuk2005,zhu2003}:
		\begin{equation} \label{eq:PertinenciaC}
		  \mu_{C,i}(t)=\left( \sum_{j=1}^{m(t)} \frac{|f(\mathbf{w}_i(t))-f_C(t)|}{|f(\mathbf{w}_j(t))-f_C(t)|} \right)^{-1},
		\end{equation} 
		\begin{equation} \label{eq:PertinenciaH}
		  \mu_{H,i}(t)=\left( \sum_{j=1}^{m(t)} \frac{|f(\mathbf{w}_i(t))-f_H(t)|}{|f(\mathbf{w}_j(t))-f_H(t)|} \right)^{-1},
		\end{equation} 
		onde $1\leq i\leq m(t)$. Assim, quando $f(\mathbf{w}_i(t))$ se aproxima de $f_C(t)$, o termo $\mu_{C,i}(t)$ se aproxima de 1, o que aproxima $\Delta \mathbf{w}_{C,i}(t)$ de 0 e, portanto, torna a influência da correlação de forças presente praticamente nula, evitando alterações no peso devidas à hegemonia presente.
		Semelhantemente, quando $f(\mathbf{w}_i(t))$ se aproxima de $f_H(t)$, o termo $\mu_{H,i}(t)$ se aproxima de 1, o que aproxima $\Delta \mathbf{w}_{H,i}(t)$ de 0, torna a influência da hegemonia histórica praticamente nula.
		\item[Crise Revolucionária:] Na etapa de crise revolucionária são executados os seguintes passos:
		\begin{enumerate}
			\item Todas as contradições $\delta_{i,j}$ são avaliadas; as contradições menores do que uma \emph{contradição mínima} $\delta_{\min}$ implicam a fusão entre os polos, de forma que
			\begin{eqnarray}
			  \delta_{i,j}(t)> \delta_{\min} \Rightarrow \mathbf{w}_i(t),\mathbf{w}_j(t) \in \Omega(t+1),\\
			  \delta_{i,j}(t)\leq \delta_{\min} \Rightarrow \mathbf{w}_i(t) \in \Omega(t+1).			  
			\end{eqnarray}
			$i\neq j$, $\forall i,j$ onde $1\leq i,j\leq m(t)$ e $\Omega(t+1)$ é o novo conjunto de polos.
			\item A partir das contradições avaliadas na etapa anterior, encontram-se aquelas maiores do que uma \emph{contradição máxima} $\delta_{\max}$; essas contradições serão con\-si\-de\-ra\-das as \emph{contradições principais} do sistema dialético, sendo os pares de polos envolvidos con\-si\-de\-ra\-dos como \emph{pares tese-antítese}, cujos \emph{polos síntese} passam também a pertencer ao novo conjunto de polos, ou seja:
			\begin{equation}
			  \delta_{i,j}(t)>\delta_{\max} \Rightarrow g(\mathbf{w}_i(t),\mathbf{w}_j(t))\in \Omega(t+1),
			\end{equation}
			para $i\neq j$, $\forall i,j$ onde $1\leq i,j\leq m(t)$.
			\item Adiciona-se o \emph{efeito de crise}, dada a \emph{máxima crise}, $\chi_{\max}$, a todos os polos do sistema dialético $\Omega(t+1)$, gerando o novo conjunto de polos, $\Omega(t+2)$, de forma que $\mathbf{w}_k(t+2)\in \Omega(t+2)$, desde que
      \begin{equation}
        w_{k,i}(t+2)=w_{k,i}(t+1)+\chi_{\max}G(0,1),
      \end{equation}
			para $1\leq k\leq m(t+1)$ e $1\leq i\leq n$, onde $G(0,1)$ é um número aleatório de distribuição gaussiana com esperança 0 e variância 1.
			\item Caso o critério de parada ainda não tenha sido atingido (número máximo de fases históricas atingido ou outro critério de parada a ser definido), é gerado um novo conjunto de polos, de forma que:
			\begin{equation}
			  \mathbf{w}_i(t+2)\in \Omega(t+2) \Rightarrow \breve{\mathbf{w}}_i(t+2)\in \Omega(t+2),
			\end{equation}
			para $1\leq i\leq m(t+2)$, onde $m(t+2)=2m(t+1)$.	Logo, o conjunto de polos é ampliado por meio da adição dos polos em antítese antagônica aos polos já existentes. Tal procedimento se dá como uma forma de modelar a concepção dialética de que, ao passar a uma nova fase história, o sistema carrega em si também os seus opostos, o que corresponde ao germe de sua potencial transformação em algo novo. Além do mais, esse procedimento repete o que foi assumido quando da inicialização do sistema dialético (geração do conjunto de polos inicial, $\Omega(0)$), podendo acelerar a convergência do algoritmo à solução ótima pela ideia de acrescentar elementos opostos à busca \cite{rahnamayan2007}. 
		\end{enumerate}
	\end{description}
\end{enumerate}

\begin{figure}
	\centering
		\includegraphics[width=0.65\textheight]{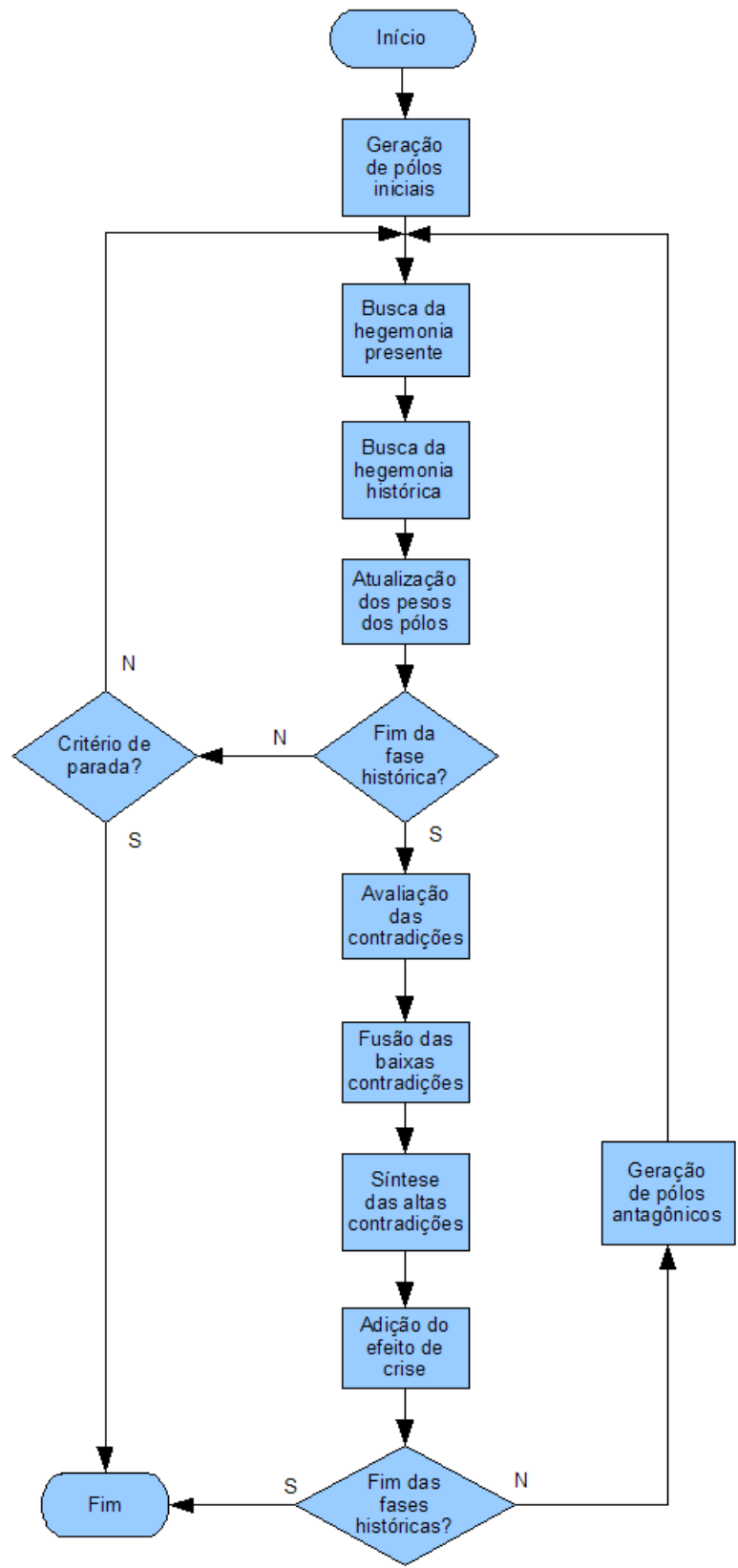}
	\caption{Fluxograma geral do ODM adaptado a busca e otimização}
	\label{fig:Fluxograma_ODM}
\end{figure}

A figura \ref{fig:Fluxograma_ODM} exibe o fluxograma geral do método dialético objetivo (ODM) adaptado a busca e otimização de forma simplificada. Embora seja relativamente mais complexo do que o fluxograma correspondente aos algoritmos de otimização por enxame de partículas (PSO), o ODM possui a propriedade de ter custo computacional decrescente em cada fase histórica, uma vez que, à medida que a solução se aproxima da solução ótima, os polos se aproximam e, portanto, as contradições entre si diminuem, o que faz com que diversos polos sejam fundidos.

\subsection{Análise pelo Método da Máxima Entropia}

Um método de busca e otimização será tão bom quanto boa for a sua capacidade de descobrir pontos no espaço de busca onde o valor da função objetivo é melhor do que o melhor antes obtido. Outra característica importante é que um algoritmo de busca e otimização será também tão bom quanto for garantida a sua convergência, que deve de preferência convergir para um ponto onde a função objetivo é ótima ou quase ótima (um máximo/mínimo global ou um máximo/mínimo local). Tem-se assim dois critérios básicos de avaliação de um algoritmo de busca e otimização:
\begin{enumerate}
  \item Capacidade de exploração, ou capacidade de descobrir de novos pontos ótimos (pontos de máximo ou de mínimo, dependendo do problema);
  \item Estabilidade e convergência.
\end{enumerate}

Diversos trabalhos buscam analisar a estabilidade e a convergência de um algoritmo de busca (algoritmos genéticos, pro\-gra\-ma\-ção evolucionária e otimização por enxame de partículas, por exemplo) usando desde critérios de estabilidade assintótica derivados do Controle de Processos \cite{jiang2007,zheng2003,trelea2003}, dado que um algoritmo pode ser considerado um processo dinâmico, até a modelagem do algoritmo de busca como uma cadeia de Markov \cite{silveira2007,poli2007,campana2006,braak2006,cantupaz2000,wright1999,suzuki1998,rudolph1994}, uma vez que o espaço de busca é finito e contável, dadas as limitações de representação numérica em sistemas computacionais, sendo o ponto de máximo/mínimo considerado um estado da cadeia.

Neste trabalho buscou-se usar uma abordagem um tanto diferente: foi utilizado o Princípio da Máxima Entropia para gerar uma versão modificada do algoritmo, que será comparada com a versão canônica do método dialético de busca e otimização aqui apresentada. A análise foi focada nos estágios de evolução, o que \emph{a priori} considera fases históricas relativamente grandes. A escolha de enfocar as etapas de evolução e não de crise revolucionária se deu baseada no fato de que a etapa de crise revolucionária não é qualitativamente diferente do que ocorre nos métodos de busca e otimização baseados em algoritmos genéticos e em programação evolucionária, a não ser pelo processo de fusão de polos semelhantes (baixa contradição entre si) e de geração de novos polos a partir das contradições principais.

Para aplicar o Princípio da Máxima Entropia no método dialético de busca e otimização é necessário partir da expressão de atualização dos pesos dos polos. Para simplificar, aqui se utilizará uma versão onde o passo histórico e o passo contemporâneo são constantes.
\begin{equation}
  \mathbf{w}_i(t+1)=\mathbf{w}_i(t) + \Delta \mathbf{w}_{C,i}(t) + \Delta \mathbf{w}_{H,i}(t), 
\end{equation}
para
\begin{equation}
  \Delta \mathbf{w}_{C,i}(t) = \eta_C(t)(1-\mu_{C,i}(t))^2 (\mathbf{w}_C(t)-\mathbf{w}_i(t)),
\end{equation}
\begin{equation}
  \Delta \mathbf{w}_{H,i}(t) = \eta_H(t)(1-\mu_{H,i}(t))^2 (\mathbf{w}_H(t)-\mathbf{w}_i(t)),
\end{equation}
\begin{equation}
  \eta_C(t)=\eta_H(t)=\eta_0,
\end{equation}
\begin{equation}
  0<\eta_0<1.
\end{equation}

Quanto à convergência do algoritmo dialético de busca, é necessário que o processo minimize as seguintes quantidades:
\begin{equation}
  E_H=\sum_{i=1}^m \mu_{H,i} |f(\mathbf{w}_i)-f_H|,
\end{equation}
\begin{equation}
  E_C=\sum_{i=1}^m \mu_{C,i} |f(\mathbf{w}_i)-f_C|.
\end{equation}

Sabendo-se que as pertinências histórica e contemporânea obedecem à seguintes propriedades:
\begin{equation}
  \sum_{i=1}^m \mu_{H,i}=1,
\end{equation}
\begin{equation}
  0\leq \mu_{H,i}\leq 1,
\end{equation}
\begin{equation}
  \sum_{i=1}^m \mu_{C,i}=1,
\end{equation}
\begin{equation}
  0\leq \mu_{C,i}\leq 1,
\end{equation}
é possível perceber aqui um caso onde as medidas de incerteza expressas pelas funções de pertinência são na verdade probabilísticas, conforme \cite{santos2009g}, cabendo portanto o uso da Entropia de Shannon, que neste caso é equivalente à Entropia de Shapley, dado que na verdade $\mu_{H,i}$ e $\mu_{C,i}$ expressam as probabilidades $p_{H,i}$ e $p_{C,i}$ de o $i$-ésimo polo coincidir com o polo hegemônico histórico e com o polo hegemônico contemporâneo, respectivamente. Assim, com base em funções de pertinência, pode-se definir uma \emph{entropia histórica} $H(\mu_H)$ e uma \emph{entropia contemporânea} $H(\mu_C)$, que podem ser maximizadas em paralelo, para maximizar a capacidade exploratória do algoritmo:
\begin{equation}
  H(\mu_H)=-\sum_{i=1}^m \mu_{H,i} \ln \mu_{H,i},
\end{equation}
\begin{equation}
  H(\mu_C)=-\sum_{i=1}^m \mu_{C,i} \ln \mu_{C,i}.
\end{equation}

Assim, o trabalho se reduz a dois problemas de maximização restritos: a) maximizar $H(\mu_H)$ para $E_H$ pequeno; b) maximizar $H(\mu_C)$ para $E_C$ pequeno. Isso se assemelha à construção de um algoritmo de agrupamento \emph{fuzzy} c-médias usando o Princípio da Máxima Entropia, utilizando o Método dos Multiplicadores de Lagrange \cite{chen2009}. Assim, são obtidas as seguintes expressões para as funções de pertinência históricas e contemporâneas \cite{santos2009g}:
\begin{equation}
  \mu_{H,i}=\frac{\exp(-\lambda_H g_{H,i})}{\sum_{j=1}^m \exp(-\lambda_H g_{H,j})}, 
\end{equation} 
\begin{equation}
  \mu_{C,i}=\frac{\exp(-\lambda_C g_{C,i})}{\sum_{j=1}^m \exp(-\lambda_C g_{C,j})}, 
\end{equation}
ou melhor:
\begin{equation} \label{eq:PertinenciaHPMEsmp}
  \mu_{H,i}=\frac{\exp(-\lambda_H|f(\mathbf{w}_i)-f_H|)}{\sum_{j=1}^m \exp(-\lambda_H |f(\mathbf{w}_j)-f_H|)}, 
\end{equation} 
\begin{equation} \label{eq:PertinenciaCPMEsmp}
  \mu_{C,i}=\frac{\exp(-\lambda_C|f(\mathbf{w}_i)-f_C|)}{\sum_{j=1}^m \exp(-\lambda_C |f(\mathbf{w}_j)-f_C|)}, 
\end{equation}
para
$$
g_{H,i}=|f(\mathbf{w}_i)-f_H|,
$$
$$
g_{C,i}=|f(\mathbf{w}_i)-f_C|,
$$
onde $\lambda_H>0$ e $\lambda_C>0$ são os multiplicadores de Lagrange para as funções de pertinência históricas e contemporâneas, respectivamente. Quando $\lambda_H\rightarrow 0$, as funções de pertinência histórica tendem a ser iguais; quando $\lambda_H\rightarrow +\infty$, a tendência é de que uma das funções de pertinência histórica tenda a quase 1, enquanto as outras tendem a 0, o que significa que um dos polos tende a convergir rapidamente para o polo hegemônico histórico \cite{rose1998,rose1990,zhang2006}. O mesmo vale para $\lambda_C$ e as funções de pertinência contemporâneas. Uma vez que os polos hegemônicos não são fixos, dado que se trata de um processo de busca, e nem a quantidade de polos é fixa, devido à etapa de crise revolucionária, as expressões das funções de pertinência se tornam:
\begin{equation} \label{eq:PertinenciaHPME}
  \mu_{H,i}(t)=\frac{\exp(-\lambda_H(t)|f(\mathbf{w}_i(t))-f_H(t)|)}{\sum_{j=1}^{m(t)} \exp(-\lambda_H(t) |f(\mathbf{w}_j(t))-f_H(t)|)}, 
\end{equation} 
\begin{equation} \label{eq:PertinenciaCPME}
  \mu_{C,i}(t)=\frac{\exp(-\lambda_C(t)|f(\mathbf{w}_i(t))-f_C(t)|)}{\sum_{j=1}^{m(t)} \exp(-\lambda_C(t) |f(\mathbf{w}_j(t))-f_C(t)|)}. 
\end{equation} 

Uma vez que $f_H(t)$ e $f_C(t)$ não são fixos, não é possível determinar algebricamente os multiplicadores de Lagrange $\lambda_H(t)$ e $\lambda_C(t)$. Uma alternativa interessante é fazer $\lambda_H(t)$ e $\lambda_C(t)$ inicialmente pequenos, para iniciar o algoritmo com uma boa capacidade de exploração, dado que as funções de pertinência passam a ter uma característica mais \emph{fuzzy} (ou seja, com transições mais suaves, por exemplo, $e^{-\lambda_1 y} > e^{-\lambda_2 y}$, para $\lambda_1 < \lambda_2$, $\forall y\in\mathbb{R}$, e $\lambda_1,\lambda_2>0$) e, à medida em que os polos são fundidos, resultado da convergência para um determinado ponto de mínimo, $\lambda_H(t)$ e $\lambda_C(t)$ aumentam. Assim, uma boa escolha poderia ser:
\begin{equation}
  \lambda_H(t)=\lambda_C(t)=\frac{1}{m(t)},
\end{equation}
o que dá origem às seguintes expressões:
\begin{equation}
  \mu_{H,i}(t)=\frac{\exp \left( -\frac{1}{m(t)} |f(\mathbf{w}_i(t))-f_H(t)| \right)}{\sum_{j=1}^{m(t)} \exp \left( -\frac{1}{m(t)} |f(\mathbf{w}_j(t))-f_H(t)| \right)}, 
\end{equation} 
\begin{equation}
  \mu_{C,i}(t)=\frac{\exp \left( -\frac{1}{m(t)} |f(\mathbf{w}_i(t))-f_C(t)| \right)}{\sum_{j=1}^{m(t)} \exp \left( -\frac{1}{m(t)} |f(\mathbf{w}_j(t))-f_C(t)| \right)}. 
\end{equation} 

Comparando as expressões \ref{eq:PertinenciaHPME} e \ref{eq:PertinenciaCPME} com as expressões canônicas \ref{eq:PertinenciaH} e \ref{eq:PertinenciaC}, nesta ordem, é possível notar o seguinte: quanto mais próximo $f(\mathbf{w}_i)$ for de $f_H$, maior a pertinência, ou seja: $\mu_{H,i} \propto 1/d(f(\mathbf{w}_i),f_H)$, onde $d:\mathbb{R}^2\rightarrow \mathbb{R}_+$ é uma medida de distância, que pode também vir a ser simplesmente o módulo da diferença, caso os argumentos sejam escalares. Assim, um caminho intuitivo é simplesmente normalizar $\mu_{H,i}=K/d(f(\mathbf{w}_i),f_H)$, onde $K\in \mathbb{R}_+$, por $\sum_{j=1}^m \mu_{H,j}=\sum_{j=1}^m K/d(f(\mathbf{w}_j),f_H)$, resultando na expressão canônica \ref{eq:PertinenciaH}, para $d(x,y)=|x-y|$, onde $x,y\in \mathbb{R}$. Logo:
$$
\lim_{f(\mathbf{w}_i)\rightarrow f_H} \mu_{H,i} = \lim_{f(\mathbf{w}_i)\rightarrow f_H} \frac{1/d(f(\mathbf{w}_i),f_H)}{\sum_{j=1}^m 1/d(f(\mathbf{w}_j),f_H)}, 
$$
que pela Regra de L'Hôpital resulta:
$$
\lim_{f(\mathbf{w}_i)\rightarrow f_H} \mu_{H,i} = \lim_{f(\mathbf{w}_i)\rightarrow f_H} \frac{1/d(f(\mathbf{w}_i),f_H)}{1/d(f(\mathbf{w}_i),f_H)} = 1.
$$

Uma outra maneira de formular o problema de obter as expressões das funções de pertinência seria fazer
$$
\mu_{H,i}\propto \exp[-d(f(\mathbf{w}_i),f_H)],
$$
o que é outra maneira de expressar que o aumento da distância entre o valor da função objetivo no $i$-ésimo polo e o valor da função no polo hegemônico histórico corresponde à diminuição da pertinência histórica do $i$-ésimo polo, evitando que o valor exploda para infinito. Assim, normaliza-se
$$
\mu_{H,i}'=K\exp[-d(f(\mathbf{w}_i),f_H)]
$$
por
$$
\sum_{j=1}^m \mu_{H,j}'=\sum_{j=1}^m K\exp[-d(f(\mathbf{w}_j),f_H)],
$$
resultando
$$
\mu_{H,i}=\frac{\mu_{H,i}'}{\sum_{j=1}^m \mu_{H,j}'},
$$
onde $K\in \mathbb{R}_+$, resultando na expressão \ref{eq:PertinenciaHPMEsmp}, para $\lambda_H=1$. Análises semelhantes podem ser feitas para as expressões das funções de pertinência contemporâneas $\mu_{C,i}$, para $1\leq i\leq m$, sendo obtidos resultados idênticos.

Logo, existem semelhanças intuitivas entre as expressões canônicas das funções de pertinência históricas e contemporâneas e aquelas expressões obtidas pelo uso do Princípio da Máxima Entropia, o que levanta a necessidade de gerar resultados experimentais que possam comparar as duas abordagens usando uma boa quantidade de funções de teste, conforme Wolpert e Macready \cite{wolpert1997}.

As diversas versões do método dialético foram comparadas com outros métodos de classificação não supervisionada e métodos evolutivos de otimização, obtendo resultados bastante razoáveis tanto quanto à tarefa de otimização, quanto ao desempenho da classificação não supervisionada, tendo sido avaliado também o custo computacional da solução \cite{santos2008a,santos2008b,santos2009a,santos2009b,santos2009c,santos2009d,santos2009e,santos2009f,santos2009g,santos2009h,santos2009i,santos2009j,santos2010a}.

\subsection{Aplicações em Classificação e Reconhecimento de Padrões} \label{sec:recpad}

Os métodos de busca e otimização baseados em computação evolucionária podem ser utilizados em diversas aplicações onde é necessário minimizar uma determinada função custo. Uma dessas aplicações é o uso de computação evolucionária para geração de métodos de agrupamento que, por sua vez, podem ser utilizados para implementar classificadores não supervisionados em aplicações como reconhecimento de padrões e classificação de imagens multiespectrais \cite{omran2005}.

Assim, o problema de agrupar os elementos do conjunto $Z=\{ \mathbf{z}_1, \mathbf{z}_2, \dots, \mathbf{z}_{n_Z} \}$, com $n_Z$ elementos, em $n_G$ grupos com centróides $V=\{ \mathbf{v}_1, \mathbf{v}_2, \dots, \mathbf{v}_{n_G} \}$ se reduz a minimizar a função
\begin{equation}
  J_e = \sum_{i=1}^{n_G} \sum_{\mathbf{z}\in G_i} \frac {d(\mathbf{z},\mathbf{v}_i)} {n_G n_{G,i}},
\end{equation} 
onde $d(\mathbf{z},\mathbf{v}_i)$ é uma medida de distância entre um vetor $\mathbf{z}$ da amostra e o centróide do $i$-ésimo grupo $\mathbf{v}_i$, podendo vir a ser a distância euclidiana, por exemplo, enquanto $n_{G,i}$ é o número de elementos de $Z$ agrupados no $i$-ésimo grupo, $G_i$, e $J_e$ é uma medida do erro de quantização \cite{omran2005,merwe2003}. Os vetores candidatos a soluções são definidos da forma que segue:
\begin{equation}
  \mathbf{x}=(\mathbf{v}_{1}^T, \mathbf{v}_{2}^T, \dots, \mathbf{v}_{n_G}^T)^T.
\end{equation}

Logo, modelando o problema para a solução utilizando o método dialético objetivo, os polos podem assumir a seguinte forma, advinda da modelagem para agrupamento e classificação não supervisionada por otimização por enxame de partículas (PSO) \cite{omran2005,merwe2003}:
\begin{equation}
  \mathbf{w}_j=(\mathbf{v}_{j,1}^T, \mathbf{v}_{j,2}^T, \dots, \mathbf{v}_{j,n_G}^T)^T,
\end{equation}
para $1\leq j\leq m$. Quanto à função a ser minimizada, evidentemente outras funções objetivo podem ser adotadas, uma vez que há diversas formas de se estimar o erro de quantização, como utilizar índices de fidelidade, por exemplo \cite{pedrini2008,omran2005}.

A escolha da função objetivo a ser minimizada depende de fatores como a velocidade do processo de agrupamento, que no caso do PSO canônico costuma ser muito mais baixa do que a de algoritmos de agrupamento como o k-médias \cite{haykin2001}, por exemplo, muito embora a qualidade do agrupamento, e portanto do resultado da classificação não supervisionada, utilizando o PSO, é significativamente maior \cite{omran2005,merwe2003}.

Também é possível montar métodos de classificação supervisionada tendo por base o conjunto de treinamento e a otimização de funções obtidas a partir da matriz de confusão, tais como a taxa de acerto global e o índice $\kappa$ de correlação estatística \cite{duda2001,duda1972}.

\subsection{Imagens Multiespectrais Sintéticas} \label{subsec:quanti_imagens}

No estudo de caso deste trabalho foram utilizadas 1086 imagens multiespectrais sintéticas de ressonância magnética com três bandas, compostas por 3258 imagens sagitais de 1~mm de espessura, resolução de 1~mm$^3$, ponderadas em densidade de prótons, $T_1$ e $T_2$, para um cérebro humano normal, obtidas por um sistema tomográfico de ressonância magnética com níveis de ruído de 0\%, 1\%, 3\%, 5\%, 7\% e 9\%, e sem inomogeneidades de campo. Essas imagens foram geradas pelo simulador de RM BrainWeb, que permite a geração de imagens sintéticas de RM ponderadas em densidade de prótons, $T_1$ e $T_2$ variando o nível de ruído e o percentual de inomogeneidade de campo \cite{brainweb1999,brainweb1998}.

\begin{figure}
	\centering
		\includegraphics[width=0.25\textwidth]{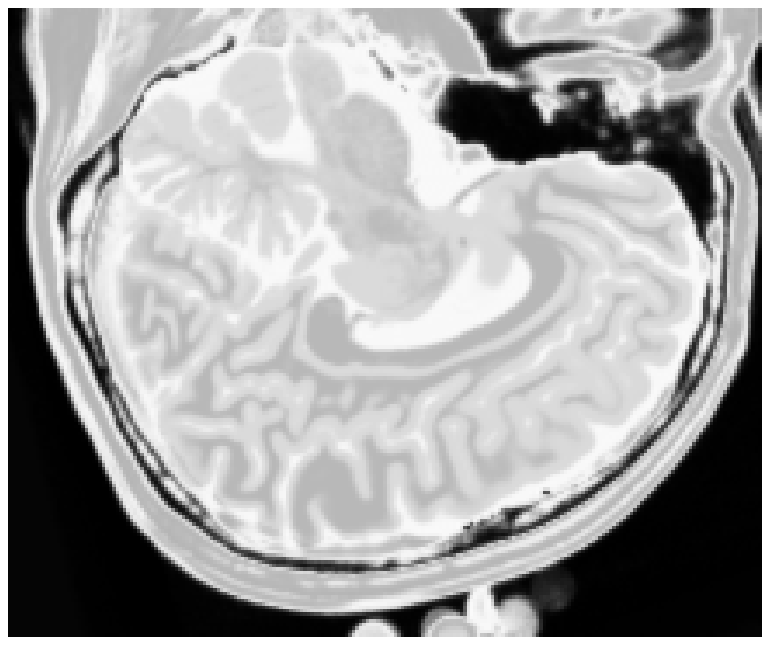}
	\caption{Imagem de RM da fatia 97 ponderada em PD}
	\label{fig:97_normal_pn0_rf0_pd}
\end{figure}

\begin{figure}
	\centering
		\includegraphics[width=0.25\textwidth]{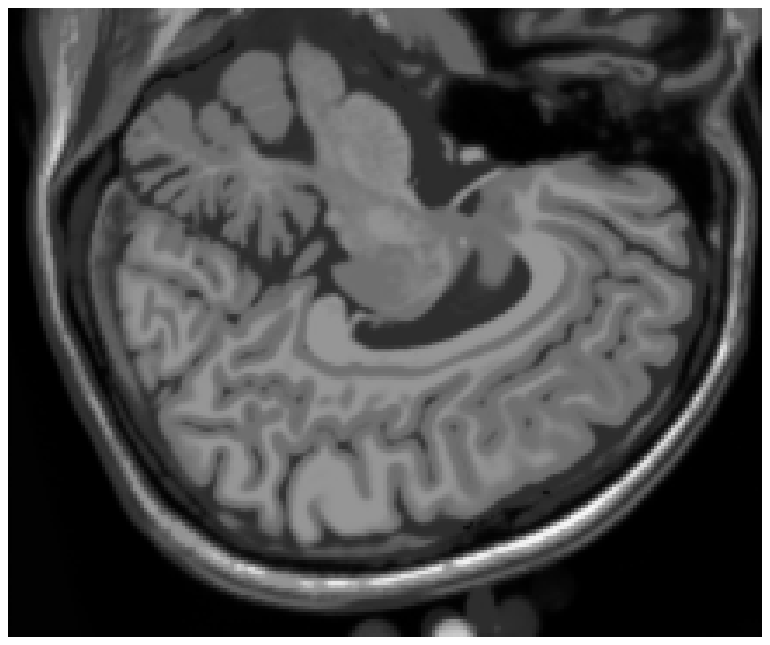}
	\caption{Imagem de RM da fatia 97 ponderada em $T_1$}
	\label{fig:97_normal_pn0_rf0_t1}
\end{figure}

\begin{figure}
	\centering
		\includegraphics[width=0.25\textwidth]{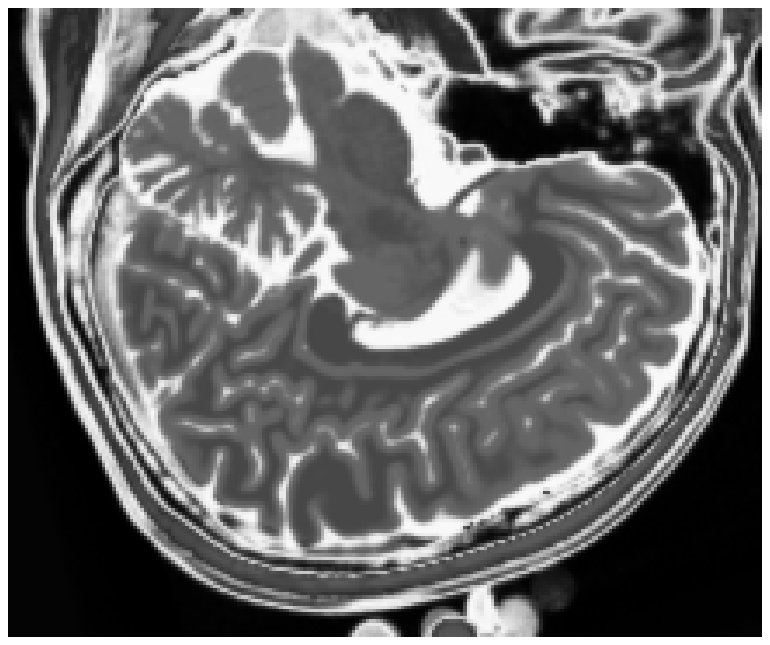}
	\caption{Imagem de RM da fatia 97 ponderada em $T_2$}
	\label{fig:97_normal_pn0_rf0_t2}
\end{figure}

\begin{figure}
	\centering
		\includegraphics[width=0.25\textwidth]{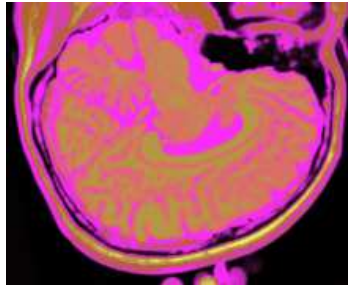}
	\caption{Composição colorida R0-G1-B2 das imagens da fatia 97 ponderadas em PD, $T_1$ e $T_2$}
	\label{fig:97_normal_pn0_rf0}
\end{figure}

As figuras \ref{fig:97_normal_pn0_rf0_pd} (banda 0), \ref{fig:97_normal_pn0_rf0_t1} (banda 1) e \ref{fig:97_normal_pn0_rf0_t2} (banda 2) mostram a fatia 97 de um volume de imagens sagitais com 181 fatias e 0\% de ruído, ponderadas em PD (densidade de próton), $T_1$ e $T_2$, enquanto a figura \ref{fig:97_normal_pn0_rf0} ilustra a composição colorida R0-G1-B2 da mesma fatia. Pode-se notar no topo do crânio, na parte inferior das imagens, a presença de artefatos, que podem ser resultantes de erros no simulador, mas que não são prejudiciais à análise, uma vez que, neste trabalho, não é dada ênfase à análise anatômica.

\subsection{Quantização Vetorial} \label{subsec:quanti_quantizacao}

A motivação da Teoria da Quantização Vetorial consiste na redução da dimensionalidade da representação dos dados de entrada, visando a compressão de dados. Assim, dado um vetor de entrada $\textbf{x}$, a função $\textbf{c}(\textbf{x})$ é dita um \emph{codificador} de $\textbf{x}$, enquanto $\textbf{x}'(\textbf{c})$ é um \emph{decodificador} de $\textbf{c}$. Na quantização vetorial, $\textbf{c}(\textbf{x})$ transforma $\textbf{x}$ de forma a poder ser representado com menos dimensões, ou por meio de uma palavra menor, enquanto $\textbf{x}'(\textbf{c})$ retorna o vetor $\textbf{c}$ à mesma base de representação de $\textbf{x}$ \cite{haykin2001}.

Ou seja, dada a transformação não linear $\Phi: \mathbb{X}\rightarrow \mathbb{A}$, onde $\textbf{x}\in \mathbb{X}$ e $\mathbb{A}$ é necessariamente um espaço de características discreto, as transformações $\textbf{c}=\Phi(\textbf{x})$ e $\textbf{x}'=\Phi^{-1}(\textbf{c})$ correspondem às tarefas de codificação e decodificação, nesta ordem, em um esquema qualquer de quantização vetorial \cite{haykin2001}.

Considerando que a entrada $\textbf{x}$ possui uma função densidade de probabilidade $p_\textbf{x}(\textbf{x})$, a \emph{distorção esperada} do sistema de quantização será dada pela expressão:
\begin{equation}
  D= \frac{1}{2} \int_{\textbf{x}\in\mathbb{X}} p_\textbf{x}(\textbf{x}) d(\textbf{x},\textbf{x}') d\textbf{x},
\end{equation}
onde $d(\textbf{x},\textbf{x}')$ é uma medida de distância entre o vetor de entrada $\textbf{x}$ e o vetor reconstruído $\textbf{x}'$, que pode ser tanto a distância euclidiana quanto outra medida de distância, como Mahalanobis, por exemplo. Uma escolha bastante comum é o quadrado da distância euclidiana: $d(\textbf{x},\textbf{x}')= ||\textbf{x}-\textbf{x}'||^2=(\textbf{x}-\textbf{x}')^T(\textbf{x}-\textbf{x}')$ \cite{haykin2001}. Assim:
\begin{equation}
  D= \frac{1}{2} \int_{\textbf{x}\in\mathbb{X}} p_\textbf{x}(\textbf{x}) ||\textbf{x}-\textbf{x}'||^2 d\textbf{x}.
\end{equation}

Um sistema de quantização é tão bom quanto menor for a distorção esperada e menor for a dimensionalidade de $\mathbb{A}$. Obviamente este é um compromisso difícil de atender, pois quanto menor a distorção, mais a dimensionalidade de $\mathbb{A}$ se aproxima da dimensionalidade de $\mathbb{X}$, e portanto menor a capacidade de compressão, ao mesmo tempo que, quanto menor a dimensionalidade de $\mathbb{A}$, maior a capacidade de compressão sobre o dado reconstruído, mas também maior é a distorção esperada.

A quantização \emph{pixel} a \emph{pixel} de imagens multiespectrais também tem um importante papel quando se deseja comprimir imagens. Na abordagem utilizada neste trabalho, a quantização de imagens multiespectrais coincide com o uso de classificadores não supervisionados baseados em centróides, ou seja, classificadores baseados em métodos de agrupamento de dados (\emph{clustering}). Assim, dado um classificador com $m$ classes tal que o conjunto de classes seja dado por:
$$
\Omega=\{ c_1, c_2, \dots, c_m \},
$$
onde a cada classe $c_j$ estão associados um vetor de pesos $\textbf{w}_j$ e uma função discriminante $g_j(\textbf{x})$, onde $1\leq j\leq m$, tem-se a seguinte regra, inspirada no critério de decisão de Bayes \cite{duda2001,duda1972}:
\begin{equation}
  g_k(\textbf{x})=\max \{g_j(\textbf{x})\}_{j=1}^n \Rightarrow \textbf{x}'=\textbf{w}_k,
\end{equation}
para $\textbf{x}=f(\textbf{u})$, onde $f:S\rightarrow [0,1]^n$ é uma imagem multiespectral normalizada de $n$ bandas, e $f':S\rightarrow [0,1]^n$ é a versão reconstruída de $f$, dado que $f'(\textbf{u})=\textbf{x}'$, $\textbf{u} \in S$ e $1\leq k\leq m$.

Como quantizadores vetoriais foram utilizados neste trabalho: mapas de k-médias, mapas auto-organizados de Kohonen com função de vizinhança gaussiana e com função de vizinhança retangular, mapas \emph{fuzzy} c-médias e classificadores dialéticos objetivos. A dimensionalidade do espaço de características codificadas $\mathbb{A}$ foi avaliada indiretamente pelo número de classes encontradas nos mapas de classificação, enquanto a distorção foi avaliada também indiretamente usando índices de fidelidade global \emph{pixel} a \emph{pixel}. A classificação não supervisionada também foi avaliada usando medidas de validade de agrupamento.

\subsection{Índices de Fidelidade}

Os principais índices de fidelidade de imagem \emph{pixel} a \emph{pixel} abordados são o \emph{erro máximo} $\epsilon_\textnormal{ME}$ (\emph{Maximum Error}, ME), o \emph{erro médio absoluto} $\epsilon_\textnormal{MAE}$ (\emph{Mean Absolute Error}, MAE), o \emph{erro médio quadrático} $\epsilon_\textnormal{MSE}$ (\emph{Mean Square Error}, MSE), a \emph{raiz do erro médio quadrático} $\epsilon_\textnormal{RMSE}$ (\emph{Root Mean Square Error}, RMSE), o \emph{erro médio quadrático normalizado} $\epsilon_\textnormal{NMSE}$ (\emph{Normalized Mean Square Error}, NMSE), a \emph{relação sinal-ruído de pico} $\epsilon_\textnormal{PSNR}$ (\emph{Peak Signal to Noise Ratio}, PSNR) e a \emph{relação sinal-ruído} $\epsilon_\textnormal{SNR}$ (\emph{Signal to Noise Ratio}, SNR), descritos da forma que segue \cite{pedrini2008}:
\begin{equation}
  \epsilon_\textnormal{ME}=\max\{||f(\textbf{u})-f'(\textbf{u})||\}_{\textbf{u}\in S},
\end{equation}
\begin{equation}
  \epsilon_\textnormal{MAE}=\frac{1}{\#S} \sum_{\textbf{u}\in S} ||f(\textbf{u})-f'(\textbf{u})||, 
\end{equation}
\begin{equation}
  \epsilon_\textnormal{MSE}=\frac{1}{\#S} \sum_{\textbf{u}\in S} ||f(\textbf{u})-f'(\textbf{u})||^2, 
\end{equation}
\begin{equation}
  \epsilon_\textnormal{RMSE}=\sqrt{\epsilon_\textnormal{MSE}}, 
\end{equation}
\begin{equation}
  \epsilon_\textnormal{NMSE}=\frac{\sum_{\textbf{u}\in S} ||f(\textbf{u})-f'(\textbf{u})||^2} {\sum_{\textbf{u}\in S} ||f(\textbf{u})||^2}, 
\end{equation}
\begin{equation}
  \epsilon_\textnormal{PSNR}=20\log_{10}\frac{L_{\max}}{\epsilon_\textnormal{RMSE}}, 
\end{equation}
\begin{equation}
  \epsilon_\textnormal{SNR}=10\log_{10}\frac{1}{\epsilon_\textnormal{NMSE}}, 
\end{equation}
para $f:S\rightarrow \{0,1,\dots,L_{\max}\}^n$. Para imagens multiespectrais normalizadas $f:S\rightarrow [0,1]^n$, tem-se $L_{\max}=1$.

Existem diversos outros índices para medição de fidelidade de imagens \cite{das2008a,das2008b,das2009}, como o índice de Wang e Bovik \cite{wang2002}. No entanto, neste trabalho o principal interesse ao usar índices de fidelidade é aproximar a medida da distorção média associada ao processo de quantização. Logo, o principal interesse está em índices baseados em diferenças \emph{pixel} a \emph{pixel}, o que não é o caso do índice de Wang e Bovik e de outros índices baseados na comparação entre estatísticas globais das imagens \cite{wang2002,pedrini2008}.

\subsection{Medidas de Validade de Agrupamento}

Os índices de validade de agrupamento (\emph{cluster validity indexes}) correspondem a funções matemáticas e estatísticas para avaliar quantitativamente um algoritmo de agrupamento. Em geral, índices de validade de agrupamento servem para determinar experimentalmente tanto o melhor número de agrupamentos (ou, usando a terminologia de classificação de imagens, o número ideal de classes presentes na imagem) quanto a melhor configuração de agrupamento, ou seja, o melhor particionamento dos dados de entrada. Neste caso, os dados de entrada são os \emph{pixels} da imagem a ser classificada ou segmentada. Assim, uma prática comum é que se façam diversos experimentos até que se encontre o melhor número de classes e a melhor configuração de particionamento \cite{omran2005,das2008a,das2008b}.

Assim, a classificação ou segmentação de uma imagem é tão boa quanto maior for a \emph{coesão} do processo de agrupamento, ou seja, quanto menor for a maior distância dos pixels ao centróide do grupo ao qual foram associados; o processo de agrupamento também tem sua qualidade avaliada pela \emph{separação} entre os centróides dos grupos: quanto maior for a menor distância entre centróides, melhor o agrupamento. Além disso, quanto menor for o \emph{erro de distorção de quantização}, melhor o processo de agrupamento \cite{das2008a,das2008b,das2009,omran2005}.

A seção \ref{sec:recpad} ilustra um índice de avaliação de agrupamento de aplicação geral, que tenta aproximar o erro de distorção, definido da forma que segue, já adaptado para o contexto de segmentação de imagens: o problema de agrupar os \emph{pixels} da imagem $f:S\rightarrow W^n$, com altura $n_H$, largura $n_W$ e $n_B=n$ bandas, resultando $n_f=n_H\times n_W$ \emph{pixels}, em $n_G$ grupos (ou classes) com centróides $V=\{ \mathbf{v}_1, \mathbf{v}_2, \dots, \mathbf{v}_{n_G} \}$ se reduz a minimizar a função
\begin{equation} \label{eq:errorquantindex}
  J_e = \sum_{i=1}^{n_G} \sum_{\forall f(\mathbf{u})\in G_i} \frac {||f(\mathbf{u})-\mathbf{v}_i||} {n_G n_{G,i}},
\end{equation} 
onde $\mathbf{u}\in S$ e $||f(\mathbf{u})-\mathbf{v}_i||$ é a distância entre o \emph{pixel} $f(\mathbf{u})$ da imagem $f:S\rightarrow W^n$ e o centróide do $i$-ésimo grupo $\mathbf{v}_i$, podendo vir a ser a distância euclidiana, por exemplo, enquanto $n_{G,i}$ é o número de elementos de $f$ agrupados no $i$-ésimo grupo e $J_e$ é uma medida do erro de quantização \cite{omran2005,merwe2003}. Os vetores candidatos a soluções são definidos da forma que segue:
\begin{equation}
  \mathbf{x}=(\mathbf{v}_{1}^T, \mathbf{v}_{2}^T, \dots, \mathbf{v}_{n_G}^T)^T.
\end{equation}

A coesão do processo de agrupamento é inversamente proporcional à distância máxima intra-agrupamento (\emph{maximum intra-clusters distance}), ou seja, a maior distância entre os \emph{pixels} e os respectivos centróides dos grupos aos quais foram alocados, sendo dada por \cite{omran2005,das2008a}:
\begin{equation}
  d_{\max}(V)=\max_{j=1,2,\dots,n_G} \left\{ \sum_{\forall f(\mathbf{u})\in G_j} \frac{||f(\mathbf{u})-\mathbf{v}_j||}{n_{G,j}} \right\}=\max_{j=1,2,\dots,n_G} S_j,
\end{equation}
onde
\begin{equation}
  S_j=\sum_{\forall f(\mathbf{u})\in G_j} \frac{||f(\mathbf{u})-\mathbf{v}_j||}{n_{G,j}},
\end{equation}
de onde vem que:
\begin{equation}
  J_e=\frac{1}{n_G}\sum_{i=1}^{n_G} S_i.
\end{equation}

A menor separação dos centróides no processo de agrupamento, ou distância mínima inter-centróides (\emph{minimum inter-clusters distance}) é importante para avaliar a presença de grupos irrelevantes, que em uma análise posterior podem ser eliminados, sendo fundidos a outros mais próximos. Esse índice é dado pela expressão \cite{omran2005,das2008a}:
\begin{equation}
  d_{\min}=\min_{\forall i,j, i\neq j}\left\{ ||\mathbf{v}_i-\mathbf{v}_j|| \right\}.
\end{equation}

Pode-se perceber que ao agrupamento ótimo correspondem a maximização da coesão e da separação, junto com a mi\-ni\-mi\-za\-ção do erro de quantização, o que equivaleria a minimizar $d_{\max}(V)$, maximizar $d_{\min}$ e minimizar $J_e$ \cite{omran2005}. Omran \emph{et al.} (2005) propõem que é possível construir um novo índice para avaliação do processo de agrupamento de imagens, que pode servir como função objetivo $f_o:W^{n\times n_G}\rightarrow W$ a ser minimizada por um algoritmo de Computação Evolucionária, a fim de obter um classificador otimizado de acordo com a minimização dessa função objetivo, tendo sido utilizado o índice $J_o$ que segue, otimizado para gerar um classificador não supervisionado k-médias \cite{omran2005}:
\begin{equation} \label{eq:omranindex}
  f_o(\mathbf{x}_i)=J_o(\mathbf{x}_i)=w_1 d_{\max}(V_i) + w_2 (L_{\max}-d_{\min}(\mathbf{x}_i)) + w_3 J_e(\mathbf{x}_i),
\end{equation}
onde $\mathbf{x}_i$ é um candidato à solução e $V_i$ é o conjunto de centróides associados ao respectivo candidato. Já as grandezas $w_1$, $w_2$ e $w_3$ são pesos, que podem ser considerados, em uma primeira abordagem, iguais, ou seja, $w_1=w_2=w_3=1/3$, não privilegiando nenhum aspecto da medida quantitativa da qualidade do agrupamento.

Há diversos outros índices de avaliação do agrupamento, tais como o índice DB, definido por \cite{das2008a}:
\begin{equation}
  \mathrm{DB}(f,V)=\frac{1}{n_G} \sum_{i=1}^{n_G} \max_{i\neq j} \left\{ \frac{S_i+S_j}{||\mathbf{v}_i-\mathbf{v}_j||}  \right\},
\end{equation}
e o índice de Xie-Beni, dado por \cite{das2009}:
\begin{equation}
  \mathrm{XB}(f,V)=\frac{\sum_{i=1}^{n_G} \sum_{\forall \mathbf{u}\in S} \mu_{i}^2(f(\mathbf{u})) ||f(\mathbf{u})-\mathbf{v}_i||^2}{n \min_{i\neq j} ||\mathbf{v}_i-\mathbf{v}_j||^2},
\end{equation}
onde
\begin{equation}
  \mu_{i}(f(\mathbf{u}))=\frac{||f(\mathbf{u})-\mathbf{v}_i||^{2/(1-q)}}{\sum_{j=1}^{n_G}  ||f(\mathbf{u})-\mathbf{v}_j||^{2/(1-q)}},
\end{equation}
para $q>1$.

Omran \emph{et al.} (2005) desenvolveram um algoritmo de agrupamento baseado na otimização do mapa de k-médias segundo a função objetivo definida na expressão \ref{eq:omranindex} utilizando a estratégia de otimização por enxame de partículas (PSO) \cite{omran2005}. Neste trabalho foram gerados resultados utilizando como função objetivo tanto o índice combinado de Omran $J_o$ quanto apenas o erro de quantização $J_e$, otimizando os parâmetros do mapa de k-médias usando o método dialético objetivo de busca e otimização.


\subsection{Método de Agrupamento de X-Médias}

O método de agrupamento de k-médias tem sido um dos métodos mais utilizados para agrupamento geral de dados, dada a sua simplicidade e baixo custo computacional \cite{haykin2001,pelleg-xmeans}. Contudo, o mapa de k-médias tem como principal inconveniente a necessidade de conhecimento prévio a respeito da quantidade de classes presente no conjunto de dados a ser agrupado \cite{pelleg-xmeans}. O mapa de x-médias é um método projetado para, dado um conjunto de dados de entrada, encontrar o número de classes que melhor agrupa os dados de entrada, obtendo-se assim um mapa de k-médias para o melhor número de classes, ótimo para um determinado índice de qualidade.

O algoritmo de x-médias é implementado de acordo com o seguinte algoritmo:
\begin{enumerate}
  \item $n_G=n_{G,\min}$
  \item Enquanto $n_G\leq n_{G,\max}$, repita:
  \begin{enumerate}
    \item Alterar parâmetros;
    \item Alterar estrutura;
  \end{enumerate}
\end{enumerate}
onde os parâmetros de entrada do algoritmo são: o número de classes $n_G$, o número mínimo $n_{G,\min}$, e o número máximo $n_{G,\max}$ de classes. O procedimento ``alterar parâmetros'' consiste em rodar o algoritmo de treinamento k-médias convencional, para um passo inicial $\eta_0$ e um máximo de $k_{\max}$ iterações. Já o procedimento ``alterar estrutura'' consiste em atualizar o número de centroides. Ao final do algoritmo, é retornado o número de classes $n_G$ que otimiza um determinado índice de qualidade $J$.

Neste trabalho foi adotada uma versão modificada do mapa de x-médias, utilizando como índice de qualidade o índice combinado de Omran, $J_o$, número de classes variando de 10 a 14, passo inicial $\eta_0=0,1$, máximo de $k_{\max}=200$ iterações. Os resultados foram obtidos usando a imagem multiespectral da fatia 97 do volume, composta pelas imagens ponderadas em densidade de prótons, $T_1$ e $T_2$, para 0\% de ruído. Os valores dos índices obtidos são mostrados na tabela \ref{tab:treinoxmedias}.
\begin{table}[htbp]
	\centering
	\footnotesize
	\caption{Índice combinado de Omran em função do número de classes, resultantes do treinamento do x-médias}
		\begin{tabular} {c|c}
			{$n_G$} & {$J_o$}\\
			\hline
			{10} & {11,47}\\
			{11} & {11,13}\\
			{12} & {10,38}\\
			{13} & {10,45}\\
			{14} & {10,45}\\
			\hline			
		\end{tabular}
	\label{tab:treinoxmedias}
\end{table}

Pode-se perceber que os resultados para 13 e 14 classes são aproximadamente iguais e correspondem aos menores valores em comparação com os restantes. Assim, pode-se escolher como número de classes presentes na imagem o menor dentre os dois, ou seja, 13 classes. O mapa de k-médias resultante foi utilizado na comparação os outros métodos de classificação não supervisionada, descritos na seção \ref{subsec:classificacao}.

\subsection{Métodos de Classificação Não Supervisionada} \label{subsec:classificacao}

Os métodos de agrupamento, como os mapas auto-organizados de Kohonen, os mapas de k-médias e \emph{fuzzy} c-médias, também podem ser utilizados como métodos de classificação não supervisionada de imagens multiespectrais. Aliás, é fácil mostrar que as tarefas de quantização vetorial e classificação multiespectral são coincidentes, a diferença é que, na primeira, o interesse está nos pesos dos centróides, enquanto na segunda, nos seus índices.

As imagens multiespectrais sintéticas obtidas pelas composições coloridas R0-G1-B2 foram classificadas usando os seguintes métodos, também utilizados para avaliar o desempenho da quantização vetorial:
\begin{enumerate}
  \item \emph{Mapa auto-organizado de Kohonen (KO)}: 3 entradas, 13 saídas, máximo de 200 iterações, taxa de aprendizado inicial $\eta_0=0,1$, arquitetura circular, função de distância gaussiana;
  \item \emph{Mapa \emph{fuzzy} c-médias (CM)}: 3 entradas, 13 saídas, máximo de 200 iterações, taxa de aprendizado inicial $\eta_0=0,1$;
  \item \emph{Mapa de k-médias (KM)}: configuração obtida da aplicação do mapa de x-médias, 3 entradas, 13 saídas, máximo de 200 iterações, taxa de aprendizado inicial $\eta_0=0,1$;
	\item \emph{Classificador dialético objetivo canônico (ODC-CAN)}: 14 polos iniciais, 2 fases históricas de 150 iterações cada fase, passo histórico inicial $\eta_0=0,1$, força mínima de $5\%$, contradição mínima de 0,01, contradição máxima de 0,98, máxima crise de $35\%$, até 12 polos. Depois de todas as fases históricas, o processo de treinamento finalizou com 13 polos;
	\item \emph{Classificador dialético objetivo com entropia máxima (ODC-PME)}: 14 polos iniciais, 2 fases históricas de 150 iterações cada fase, passo histórico inicial $\eta_0=0,1$, força mínima de $5\%$, contradição mínima de 0,01, contradição máxima de 0,98, máxima crise de $35\%$, até 12 polos. Depois de todas as fases históricas, o processo de treinamento finalizou com 13 polos. Aqui foi aplicado o Princípio da Máxima Entropia para obter novas expressões para as funções de anticontradição, semelhantemente ao \emph{fuzzy} c-médias com máxima entropia, resultando praticamente as mesmas expressões das funções de pertinência do \emph{fuzzy} c-médias com máxima entropia \cite{chen2009}.
\end{enumerate}

O número de entradas é igual ao número de bandas da imagem, uma vez que a classificação é \emph{pixel} a \emph{pixel}, sendo portanto o \emph{pixel} a entrada da rede. Já o número de saídas foi definido como sendo o número de classes presentes na imagem, que por sua vez é o número de elementos anatômica e fisiologicamente diferentes presentes na imagem (11 classes) mais duas classes para alocar diferentes populações de ruído \cite{brainweb1998,brainweb1999}. Os demais parâmetros foram definidos empiricamente. Note que o número final de polos dos classificadores dialéticos coincidiu com o número de classes obtido pelo mapa de x-médias: 13 classes.

\subsection{Classificadores Otimizados} \label{subsec:classificacaootimizada}

As imagens multiespectrais sintéticas obtidas pelas composições coloridas R0-G1-B2 foram classificadas usando os seguintes métodos baseados na otimização do mapa de k-médias, considerando 13 classes, 20 polos iniciais, máximo de 10 fases históricas, com duração máxima de 20 gerações cada fase histórica, passos iniciais (histórico e contemporâneo) de 0,99, contradição mínima de 0,1, contradição máxima de 0,9, efeito de crise máximo de 0,9, valor limite de 0,01:
\begin{enumerate}
  \item \emph{EQ-CAN-KM}: k-médias otimizado pelo método dialético canônico, de acordo com o erro de quantização $J_e$;
  \item \emph{IC-CAN-KM}: k-médias otimizado pelo método dialético canônico, de acordo com o índice combinado de Omran $J_o$;
  \item \emph{EQ-PME-KM}: k-médias otimizado pelo método dialético com entropia maximizada, de acordo com o erro de quantização $J_e$;
  \item \emph{IC-PME-KM}: k-médias otimizado pelo método dialético com entropia maximizada, de acordo com o índice combinado de Omran $J_o$.
\end{enumerate}

Os parâmetros iniciais foram definidos de forma semelhante ao que foi descrito na seção anterior. Os parâmetros do método dialético foram definidos empiricamente.

Uma vez que a velocidade de convergência e os resultados de qualquer algoritmo de busca e otimização são altamente dependentes da inicialização do algoritmo, foram utilizadas 5 iterações do algoritmo k-médias padrão para geração do conjunto inicial de polos, de acordo com o critério de geração da população inicial proposto por Saha e Bandyopadhyay (2005), que evita buscas iniciais totalmente aleatórias, uma vez que, embora apenas 5 iterações não sejam suficientes para garantir a convergência do algoritmo k-médias, e nem essa convergência é desejada, uma vez que se está utilizando um algoritmo de otimização em substituição ao algoritmo de treinamento, essas 5 iterações iniciais ajudam o algoritmo de otimização a iniciar com polos que podem estar na vizinhança da solução ótima, sendo este portanto um critério melhor do que inicializar aleatoriamente, evitando buscas totalmente aleatórias e variações quanto à convergência do algoritmo \cite{saha2008}.

\section{Resultados e Discussões} \label{sec:quanti_resultados}

Para facilitar a avaliação dos resultados obtidos, estes foram divididos em duas categorias: resultados gerados usando classificadores dialéticos objetivos; e resultados gerados usando mapa de k-médias otimizado pelo método dialético de otimização, em suas versões canônica e com maximização da entropia, em função do índice combinado de Omran e do erro de quantização. Tanto a qualidade da quantização quanto a da segmentação são avaliadas usando índices de validade de agrupamento e índices de fidelidade da imagem quantizada em relação à imagem original.

\subsection{Classificadores Dialéticos Objetivos}

\begin{figure}
  \centering
  \begin{minipage}[b]{0.48\linewidth}
    \centering
    \includegraphics[width=0.5\linewidth]{97_normal_pn0_rf0.eps}
    \\(a)\\
	  \includegraphics[width=0.5\linewidth]{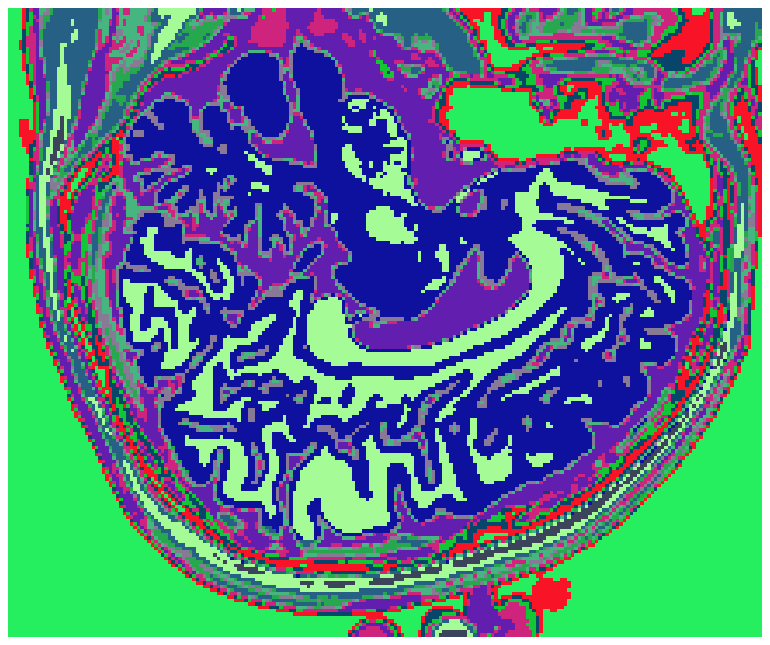}
    \\(b)\\
	  \includegraphics[width=0.5\linewidth]{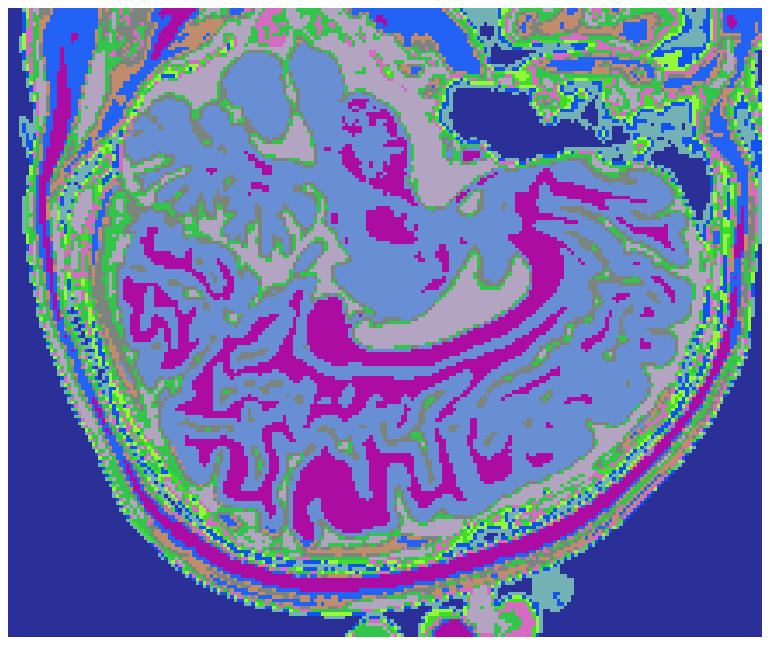}
    \\(c)\\
  \end{minipage}
  \begin{minipage}[b]{0.48\linewidth}
    \centering
    \includegraphics[width=0.5\linewidth]{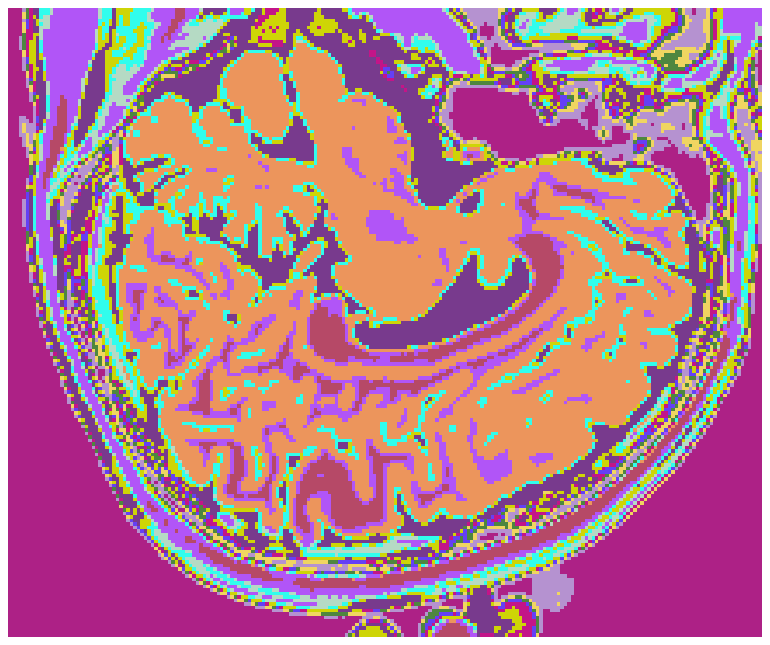}
    \\(d)\\
	  \includegraphics[width=0.5\linewidth]{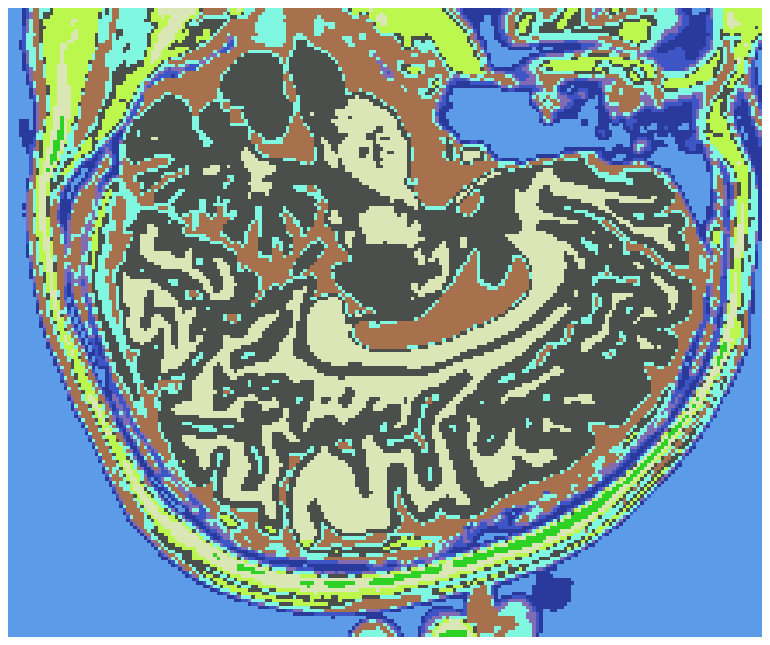}
    \\(e)\\
	  \includegraphics[width=0.5\linewidth]{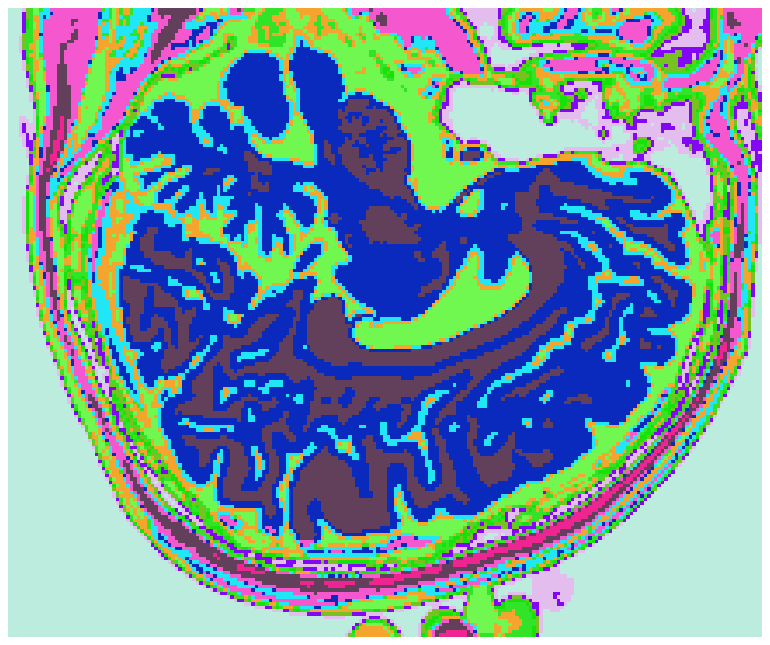}
    \\(f)\\
  \end{minipage}
  \caption{Composição colorida R0-G1-B2 das imagens da fatia 97 ponderadas em PD, $T_1$ e $T_2$ (a) e resultados de classificação usando os métodos KO (b), CM (c), KM (d), ODC-PME (e) e ODC-CAN (f)}
  \label{fig:resultados_classificacoesODC}
\end{figure}

\begin{figure}
  \centering
  \begin{minipage}[b]{0.48\linewidth}
    \centering
    \includegraphics[width=0.5\linewidth]{97_normal_pn0_rf0.eps}
    \\(a)\\
	  \includegraphics[width=0.5\linewidth]{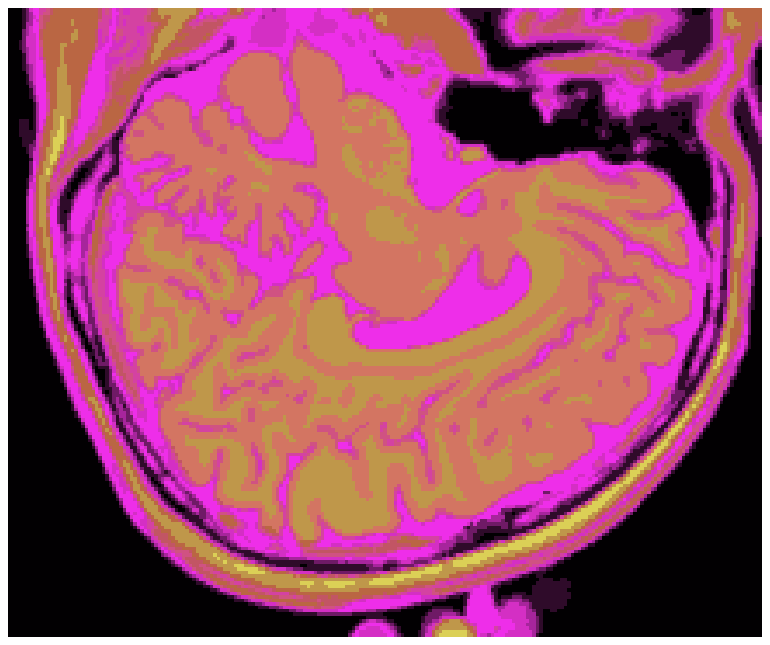}
    \\(b)\\
	  \includegraphics[width=0.5\linewidth]{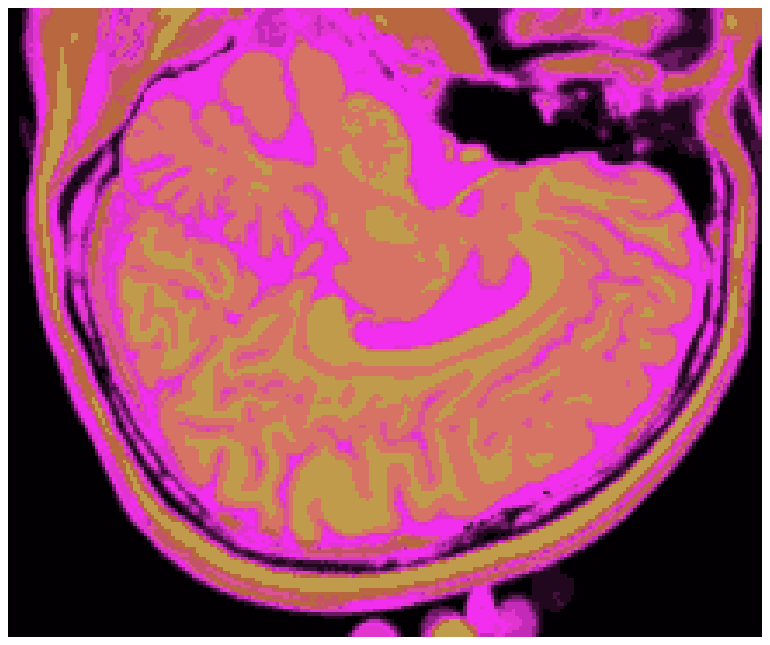}
    \\(c)\\
  \end{minipage}
  \begin{minipage}[b]{0.48\linewidth}
    \centering
    \includegraphics[width=0.5\linewidth]{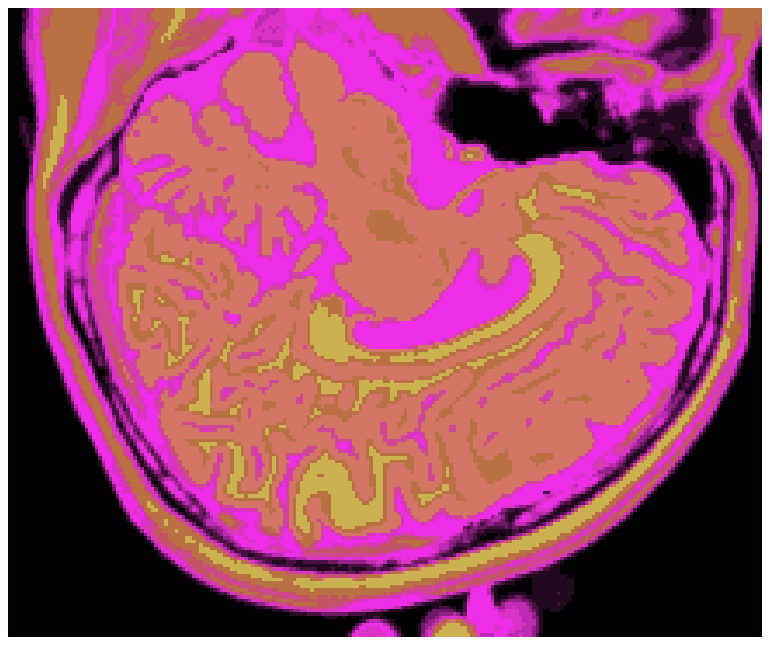}
    \\(d)\\
	  \includegraphics[width=0.5\linewidth]{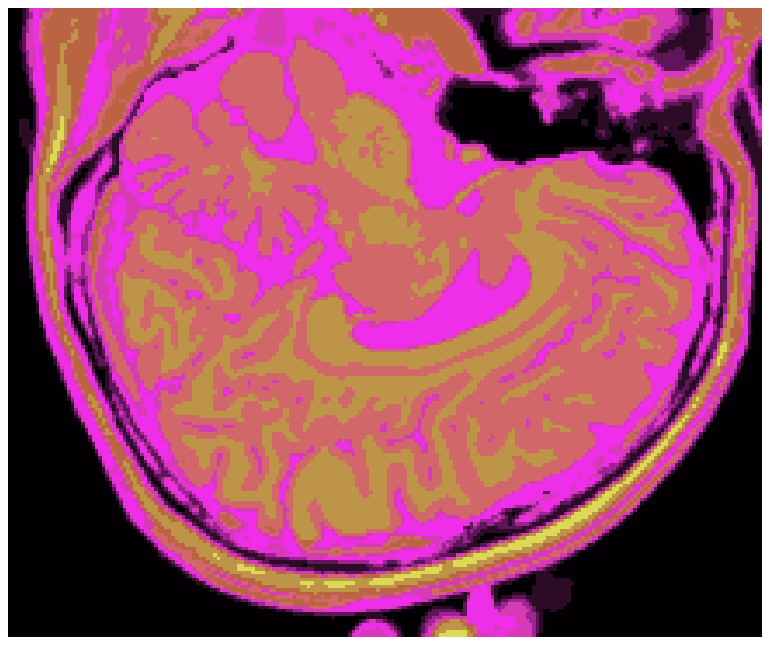}
    \\(e)\\
	  \includegraphics[width=0.5\linewidth]{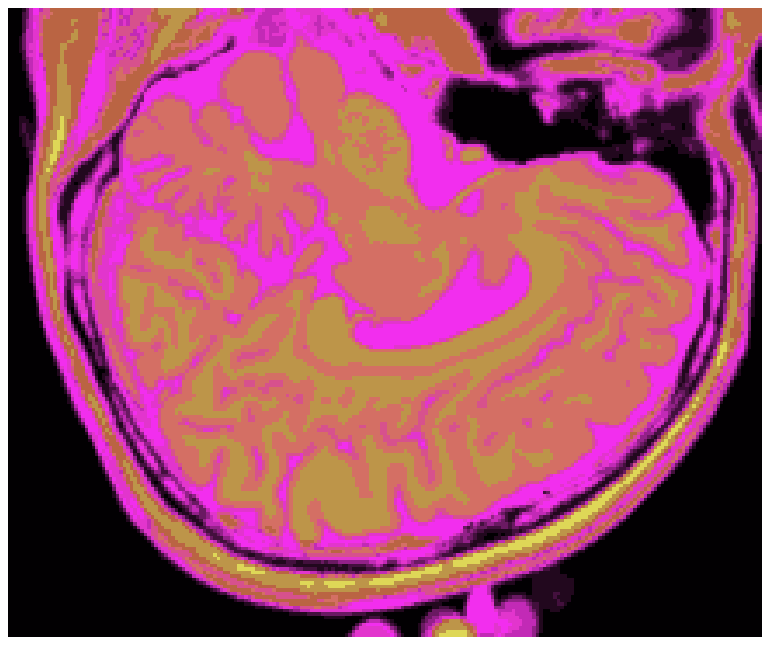}
    \\(f)\\
  \end{minipage}
  \caption{Composição colorida R0-G1-B2 das imagens da fatia 97 ponderadas em PD, $T_1$ e $T_2$ (a) e resultados de quantização usando os métodos KO (b), CM (c), KM (d), ODC-PME (e) e ODC-CAN (f)}
  \label{fig:resultados_quantizacoesODC}
\end{figure}

A figura \ref{fig:resultados_classificacoesODC} mostra os resultados de classificação, enquanto a figura \ref{fig:resultados_quantizacoesODC} exibe os resultados de quantização para a imagem sem ruído da fatia 97, figura \ref{fig:97_normal_pn0_rf0}, usando os métodos KO, CM, KM, ODC-PME e ODC-CAN. Esses resultados ilustram qualitativamente as diferenças entre os métodos de classificação e quantização, dado que a fatia 97 possui todas as 13 classes presentes na análise \cite{brainweb1999}.

\begin{figure}[ht]
	\centering
		\includegraphics[width=0.7\textwidth]{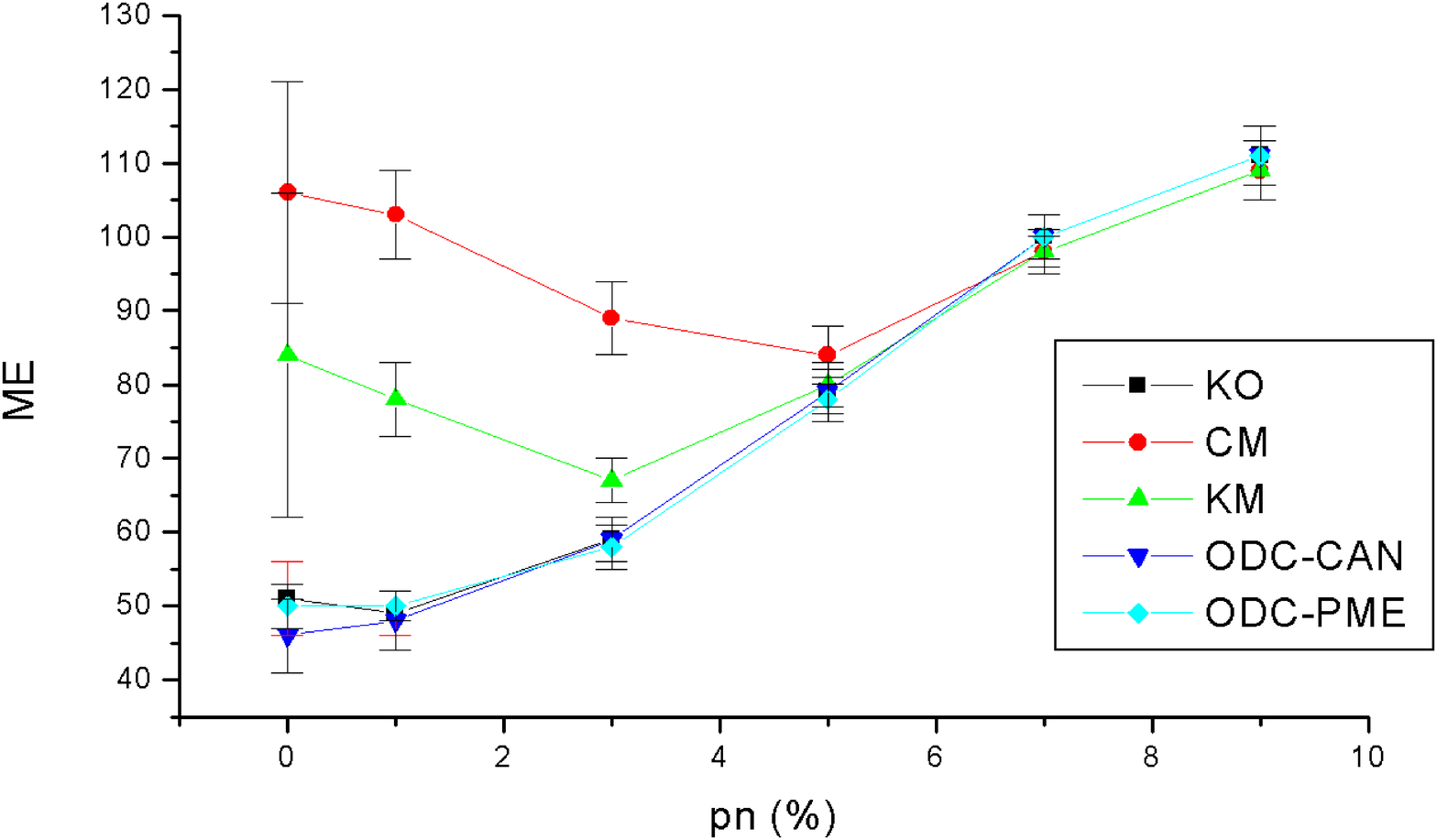}
	\caption{Resultados de $\epsilon_\textnormal{ME}$ em função do ruído percentual (ODC)}
	\label{fig:ME}
\end{figure}

\begin{figure}[ht]
	\centering
		\includegraphics[width=0.7\textwidth]{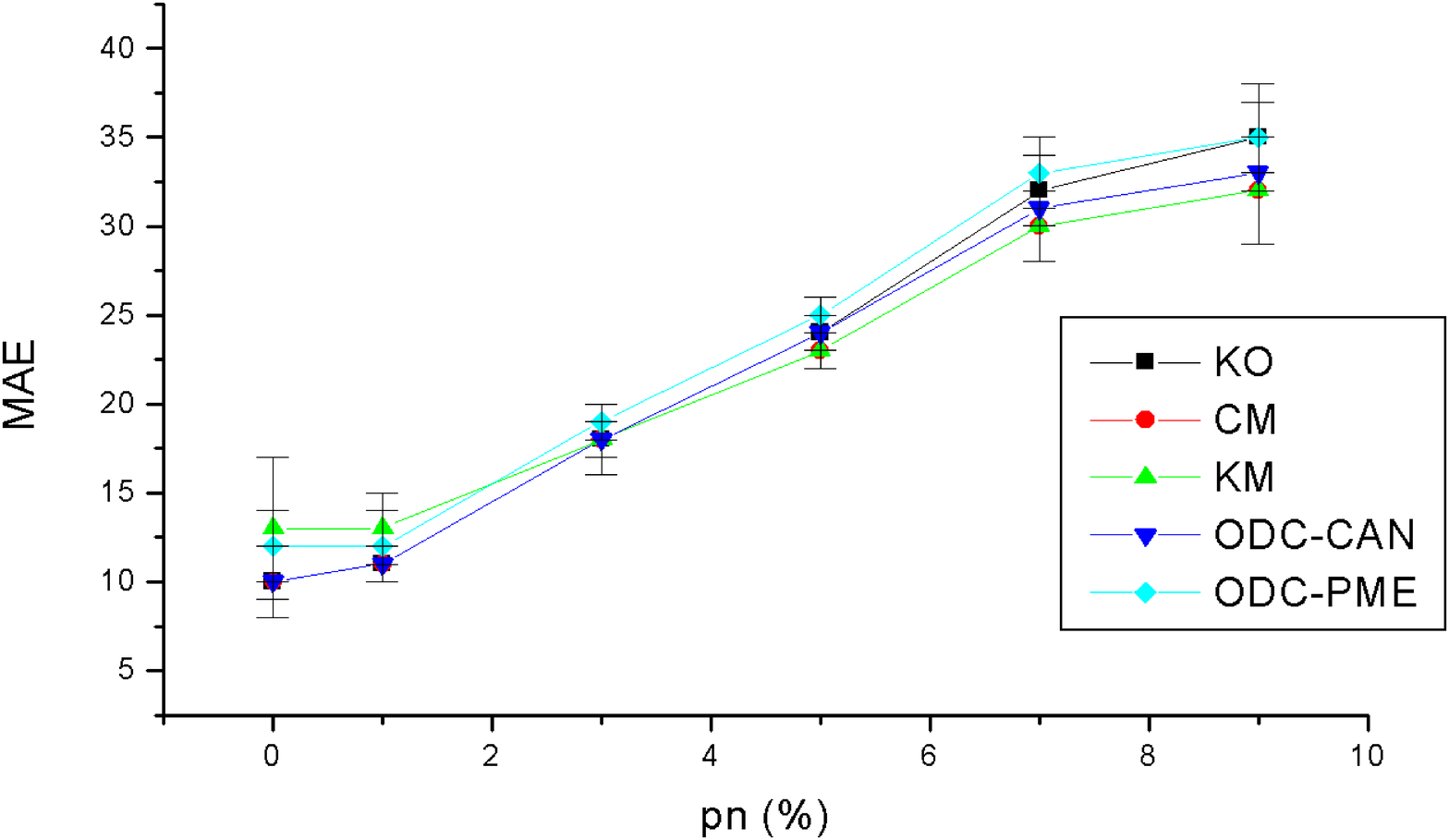}
	\caption{Resultados de $\epsilon_\textnormal{MAE}$ em função do ruído percentual (ODC)}
	\label{fig:MAE}
\end{figure}

\begin{figure}[ht]
	\centering
		\includegraphics[width=0.7\textwidth]{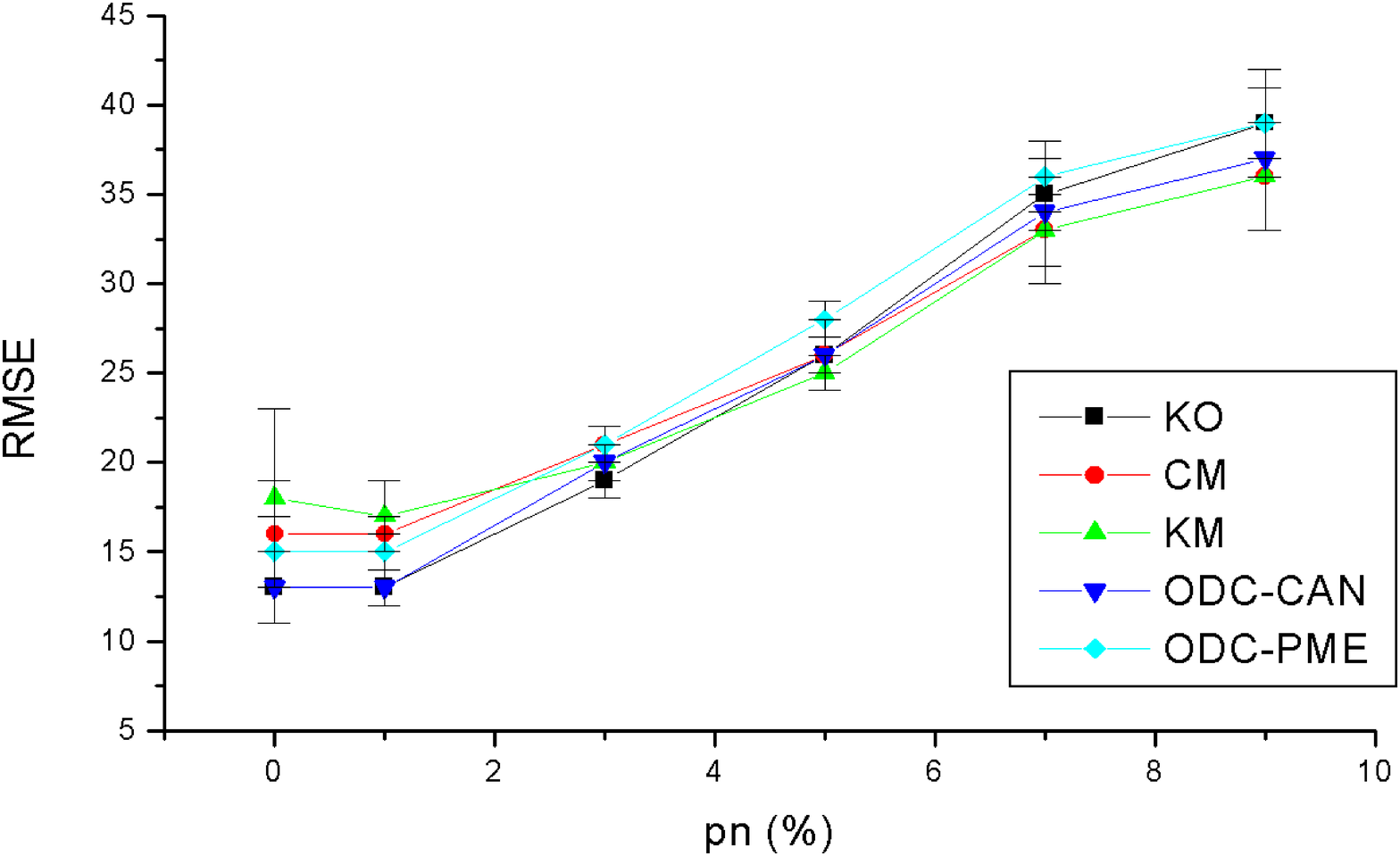}
	\caption{Resultados de $\epsilon_\textnormal{RMSE}$ em função do ruído percentual (ODC)}
	\label{fig:RMSE}
\end{figure}

\begin{figure}[ht]
	\centering
		\includegraphics[width=0.7\textwidth]{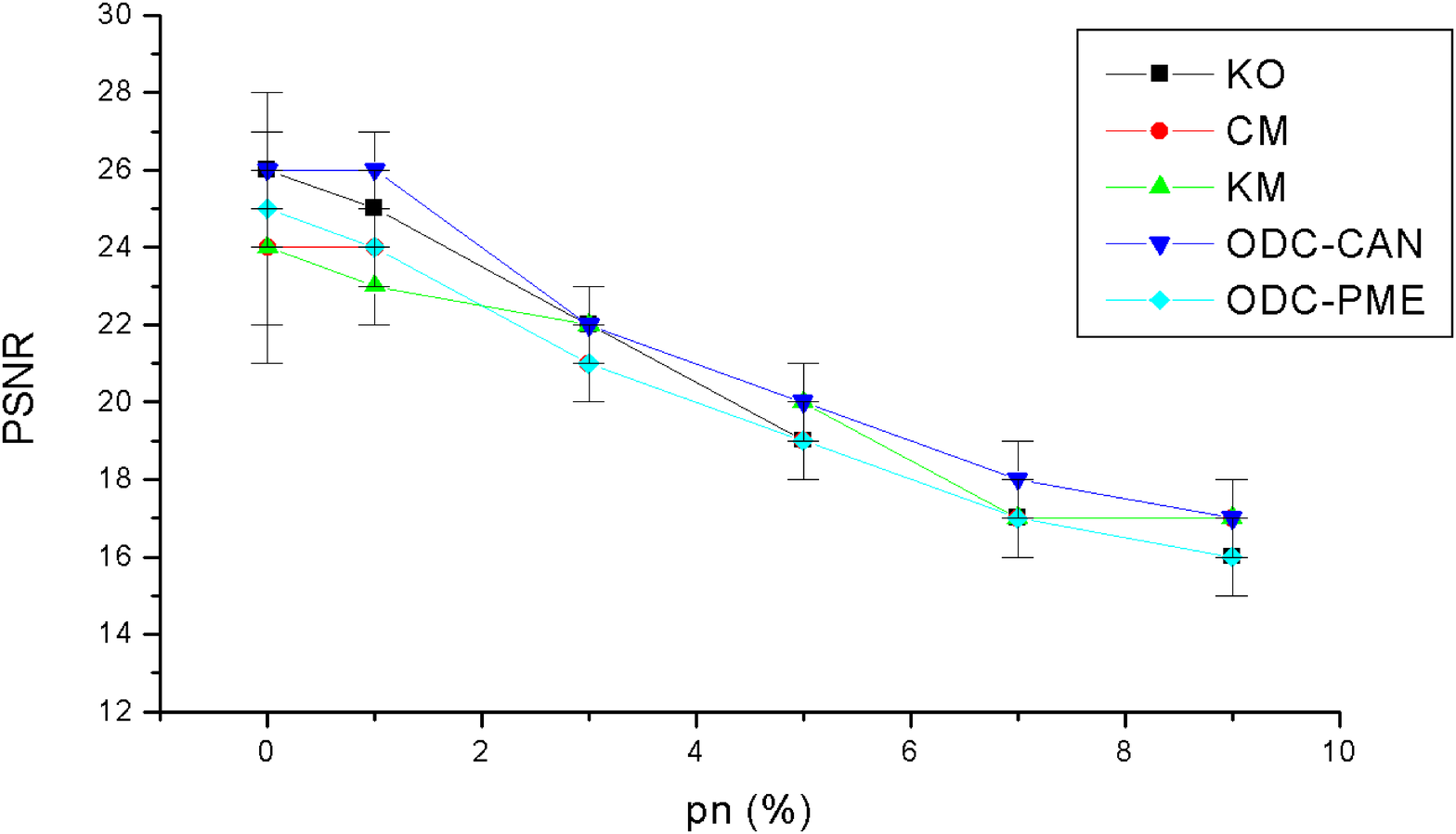}
	\caption{Resultados de $\epsilon_\textnormal{PSNR}$ em função do ruído percentual (ODC)}
	\label{fig:PSNR}
\end{figure}

A tabela \ref{tab:quant} mostra os resultados da avaliação dos métodos de classificação não supervisionada quanto à quantização vetorial, usando os índices de fidelidade $\epsilon_\textnormal{ME}$, $\epsilon_\textnormal{MAE}$, $\epsilon_\textnormal{MSE}$, $\epsilon_\textnormal{RMSE}$ e $\epsilon_\textnormal{PSNR}$, considerando todas as 181 fatias com 3 bandas (DP, $T_1$ and $T_2$), para os métodos KO, CM, KM, ODC-PME e ODC-CAN e 0\% de ruído. Já os gráficos das figuras \ref{fig:ME}, \ref{fig:MAE}, \ref{fig:RMSE} e \ref{fig:PSNR} mostram os resultados em função do nível percentual de ruído para os diversos métodos, para um total de 6 volumes de 181 fatias de 3 bandas, totalizando 1086 imagens coloridas, ou 3258 imagens em níveis de cinza. 


\begin{figure}[ht]
	\centering
		\includegraphics[width=0.7\textwidth]{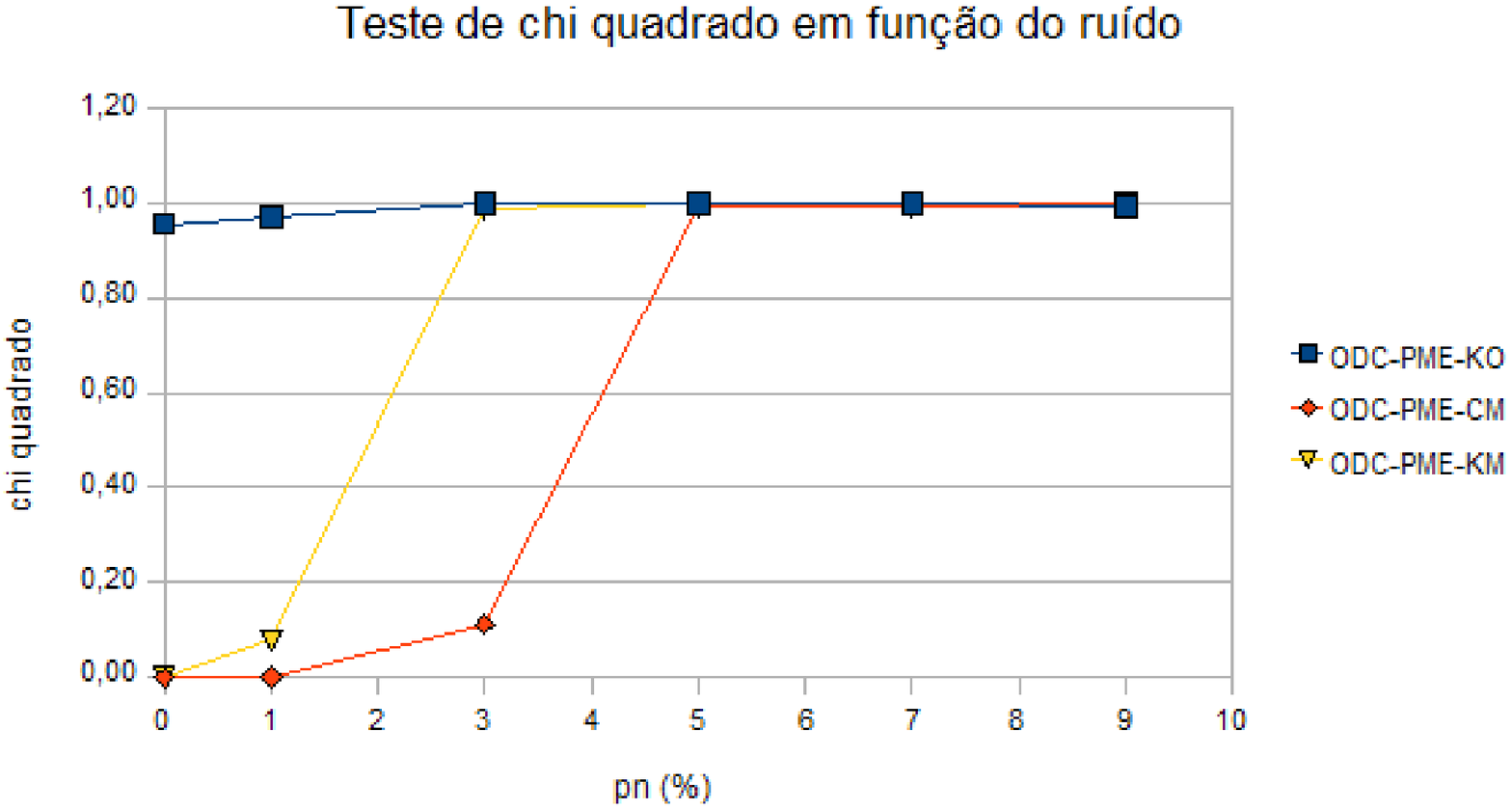}
	\caption{Graus de similaridade do método ODC-PME em relação aos métodos KO, CM e KM, usando o teste de $\chi^2$, em função do ruído percentual}
	\label{fig:chi2PME}
\end{figure}

\begin{figure}[ht]
	\centering
		\includegraphics[width=0.7\textwidth]{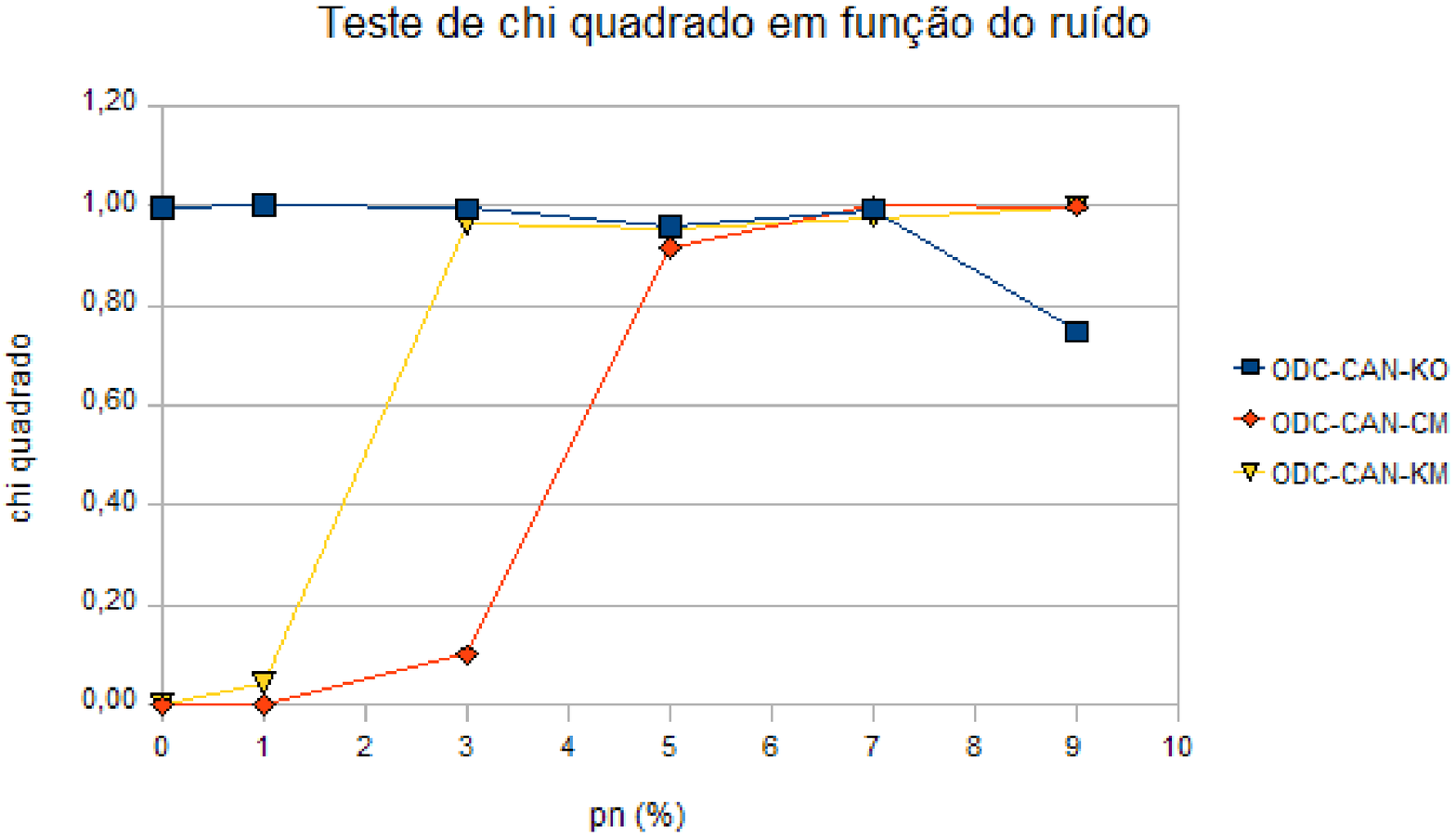}
	\caption{Graus de similaridade do método ODC-CAN em relação aos métodos KO, CM e KM, usando o teste de $\chi^2$, em função do ruído percentual}
	\label{fig:chi2CAN}
\end{figure}

As tabelas \ref{tab:individual_simPME} e \ref{tab:individual_simCAN} mostram o resultado das comparações entre os pares de média e desvio médio das medidas de $\epsilon_\textnormal{ME}$, $\epsilon_\textnormal{MAE}$, $\epsilon_\textnormal{MSE}$, $\epsilon_\textnormal{RMSE}$ e $\epsilon_\textnormal{PSNR}$, para os métodos KO, CM, KM, ODC-CAN e ODC-PME, de acordo com teste F de hipótese nula, enquanto as tabelas \ref{tab:global_simPME} e \ref{tab:global_simCAN} mostram comparações entre os métodos ODC-PME e ODC-CAN, e os métodos KO, CM e KM, respectivamente, usando o teste de $\chi^2$ para medir o grau de dependência (considerado aqui como grau de similaridade) entre os resultados dos métodos ODC-PME e ODC-CAN e os resultados dos outros métodos; as sequências de valores observados foram construídas usando os pares de média e desvio médio dos índices de fidelidade $\epsilon_\textnormal{ME}$, $\epsilon_\textnormal{MAE}$, $\epsilon_\textnormal{RMSE}$ e $\epsilon_\textnormal{PSNR}$. Todos os testes foram executados considerando um intervalo de confiança de 95\%. As figuras \ref{fig:chi2PME} e \ref{fig:chi2CAN} mostram o resultado da similaridade medida entre os métodos ODC-PME e ODC-CAN e os outros métodos, respectivamente, usando o teste de $\chi^2$, em função do ruído percentual.
\begin{table}[htbp]
	\centering
	\footnotesize
	\caption{Resultados de quantização expressos pelos índices de fidelidade, para 0\% de ruído}
		\begin{tabular} {c|c|c|c|c|c}
			{} & {KO} & {CM} & {KM} & {ODC-PME} & {ODC-CAN}\\
			\hline
			{$\epsilon_\textnormal{ME}$} & {$50\pm 5$} & {$106\pm 15$} & {$84\pm 22$} & {$50\pm 3$} & {$46\pm 6$}\\
			{$\epsilon_\textnormal{MAE}$} & {$10\pm 3$} & {$10\pm 2$} & {$13\pm 4$} & {$12\pm 2$} & {$10\pm 2$}\\
			{$\epsilon_\textnormal{MSE}$} & {$186\pm 52$} & {$258\pm 81$} & {$347\pm 261$} & {$249\pm 60$} & {$169\pm 46$}\\
			{$\epsilon_\textnormal{RMSE}$} & {$13\pm 2$} & {$16\pm 31$} & {$18\pm 5$} & {$15\pm 2$} & {$13\pm 2$}\\
			{$\epsilon_\textnormal{PSNR}$} & {$26\pm 2$} & {$24\pm 2$} & {$24\pm 3$} & {$25\pm 1$} & {$26\pm 2$}\\
			\hline			
		\end{tabular}
	\label{tab:quant}
\end{table}
\begin{table}[htbp]
	\centering
	\footnotesize
	\caption{Graus de similaridade entre o método ODC-PME e os métodos KO, CM e KM, de acordo com o teste de hipótese nula F, para 0\% de ruído}
		\begin{tabular} {c|c|c|c|c}
			{} & {$\mu(\epsilon_\textnormal{ME})$} & {$\mu(\epsilon_\textnormal{MAE})$} & {$\mu(\epsilon_\textnormal{RMSE})$} & {$\mu(\epsilon_\textnormal{PSNR})$}\\
			\hline
			{ODC-PME-KO} & {$0,99$} & {$0,86$} & {$0,89$} & {$1,00$}\\
			{ODC-PME-CM} & {$0,61$} & {$0,86$} & {$0,89$} & {$1,00$}\\
			{ODC-PME-KM} & {$0,83$} & {$0,86$} & {$0,89$} & {$1,00$}\\
			\hline			
		\end{tabular}
	\label{tab:individual_simPME}
\end{table}
\begin{table}[htbp]
	\centering
	\footnotesize
	\caption{Graus de similaridade entre o método ODC-PME e os métodos KO, CM e KM, de acordo com o teste de $\chi^2$, em função do ruído}
		\begin{tabular} {c|c|c|c}
			{pn (\%)} & {ODC-PME-KO} & {ODC-PME-CM} & {ODC-PME-KM}\\
			\hline
			{0} & {0,956111} & {0,000000} & {0,000012}\\
			{1} & {0,970435} & {0,000060} & {0,079578}\\
			{3} & {0,999864} & {0,112552} & {0,986766}\\
			{5} & {0,999971} & {0,994726} & {0,998794}\\
			{7} & {1,000000} & {0,995641} & {0,992783}\\
			{9} & {0,994829} & {0,998837} & {0,998837}\\
			\hline			
		\end{tabular}
	\label{tab:global_simPME}
\end{table}
\begin{table}[htbp]
	\centering
	\footnotesize
	\caption{Graus de similaridade entre o método ODC-CAN e os métodos KO, CM e KM, de acordo com o teste de hipótese nula F, para 0\% de ruído}
		\begin{tabular} {c|c|c|c|c}
			{} & {$\mu(\epsilon_\textnormal{ME})$} & {$\mu(\epsilon_\textnormal{MAE})$} & {$\mu(\epsilon_\textnormal{RMSE})$} & {$\mu(\epsilon_\textnormal{PSNR})$}\\
			\hline
			{ODC-CAN-KO} & {$0,93$} & {$1,00$} & {$1,00$} & {$0,97$}\\
			{ODC-CAN-CM} & {$0,54$} & {$1,00$} & {$1,00$} & {$0,97$}\\
			{ODC-CAN-KM} & {$0,74$} & {$1,00$} & {$1,00$} & {$0,97$}\\
			\hline			
		\end{tabular}
	\label{tab:individual_simCAN}
\end{table}
\begin{table}[htbp]
	\centering
	\footnotesize
	\caption{Graus de similaridade entre o método ODC-CAN e os métodos KO, CM e KM, de acordo com o teste de $\chi^2$, em função do ruído}
		\begin{tabular} {c|c|c|c}
			{pn (\%)} & {ODC-CAN-KO} & {ODC-CAN-CM} & {ODC-CAN-KM}\\
			\hline
			{0} & {0,994985} & {0,000000} & {0,000005}\\
			{1} & {0,999750} & {0,000069} & {0,044842}\\
			{3} & {0,993934} & {0,100309} & {0,962277}\\
			{5} & {0,956870} & {0,915891} & {0,954338}\\
			{7} & {0,992681} & {0,998602} & {0,976113}\\
			{9} & {0,747098} & {0,997809} & {0,997809}\\
			\hline			
		\end{tabular}
	\label{tab:global_simCAN}
\end{table}


Observando o gráfico da figura \ref{fig:ME} pode-se notar que os métodos ODC-CAN, ODC-PME e KO são os que atingem os melhores resultados para $\epsilon_\textnormal{ME}$, para 0\% de ruído. No entanto, os resultados vão se aproximando à medida em que o nível de ruído aumenta, e acabam por se tornar totalmente coincidentes a partir de 5\% de ruído. Já os gráficos das figuras \ref{fig:MAE}, \ref{fig:RMSE} e \ref{fig:PSNR} mostram que os resultados de $\epsilon_\textnormal{MAE}$, $\epsilon_\textnormal{RMSE}$ e $\epsilon_\textnormal{PSNR}$ obtidos pelos métodos de classificação utilizados não são distinguíveis, não sendo possível comparar os métodos entre si usando os índices de fidelidade citados. Consequentemente, a comparação dos métodos índice a índice não é muito conclusiva quanto à classificação dos métodos de segmentação, como mostram os gráficos das figuras \ref{fig:ME}, \ref{fig:MAE}, \ref{fig:RMSE} e \ref{fig:PSNR}, para todos os níveis de ruído, e os testes F exibidos nas tabelas \ref{tab:individual_simPME} e \ref{tab:individual_simCAN}, que comparam os métodos ODC-PME e ODC-CAN com os métodos KO, CM e KM, para 0\% de ruído.

Uma vez que as comparações índice a índice não são muito conclusivas, foi utilizado o teste de $\chi^2$ para fazer comparações globais entre os métodos ODC-PME e ODC-CAN e os métodos KO, CM e KM. A tabela \ref{tab:global_simPME} mostra os resultados do teste de $\chi^2$ entre o método ODC-PME e os outros métodos, mostrando que, para 0\% de ruído, os métodos ODC-PME e KO são idênticos. O gráfico da figura \ref{fig:chi2PME} ilustra os resultados para os níveis de ruído de 0\% a 9\%, mostrando que os métodos ODC-PME e KO são praticamente iguais quanto aos resultados dos índices de fidelidade, apesar de todos os métodos atingirem resultados idênticos a partir de 5\% de ruído.

A tabela \ref{tab:global_simCAN} mostra os resultados do teste de $\chi^2$ entre o método ODC-CAN e os outros métodos, mostrando que, para 0\% de ruído, os métodos ODC-CAN e KO são similares. O gráfico da figura \ref{fig:chi2CAN} ilustra os resultados para os níveis de ruído de 0\% a 9\%, mostrando que os métodos ODC-CAN e KO são muito similares quanto aos resultados dos índices de fidelidade até 7\% de ruído, apesar de todos os métodos atingirem resultados semelhantes a partir de 5\% de ruído, enquanto a partir de 7\% de ruído os métodos ODC-CAN e KO passam a se diferenciar, comportamento bem diferente daquele observado no método ODC-PME quando comparado com o método KO.

Esses resultados mostram os classificadores dialéticos objetivos em sua versão canônica e na versão com entropia maximizada podem atingir resultados de segmentação e quantização similares àqueles obtidos usando mapas auto-organizados de Kohonen, principalmente quando se trata da versão com entropia maximizada, o que é uma característica muito importante, uma vez que os mapas auto-organizados de Kohonen são quantizadores vetoriais ótimos \cite{haykin2001}, com a vantagem de os classificadores dialéticos objetivos terem número de classes adaptável, não sendo necessário conhecer o número de classes presente na imagem, bastando inicializar a classificação com um número de classes maior do que o esperado.

\subsection{Classificadores K-Médias Otimizados pelo Método Dialético}

\begin{figure}
	\centering
		\includegraphics[width=0.7\textwidth]{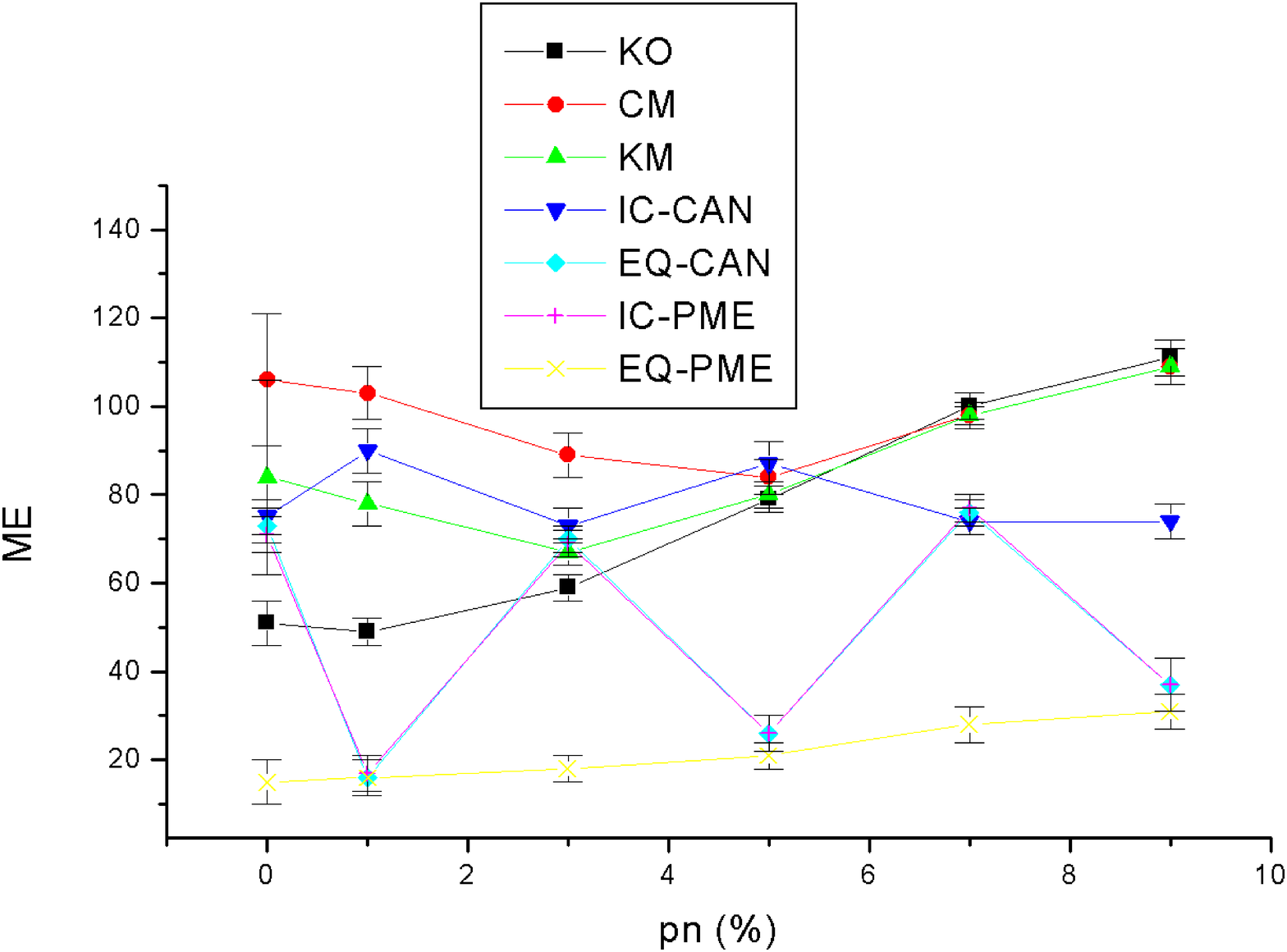}
	\caption{Resultados de $\epsilon_\textnormal{ME}$ em função do ruído percentual (ODM)}
	\label{fig:MEotm}
\end{figure}

\begin{figure}
	\centering
		\includegraphics[width=0.7\textwidth]{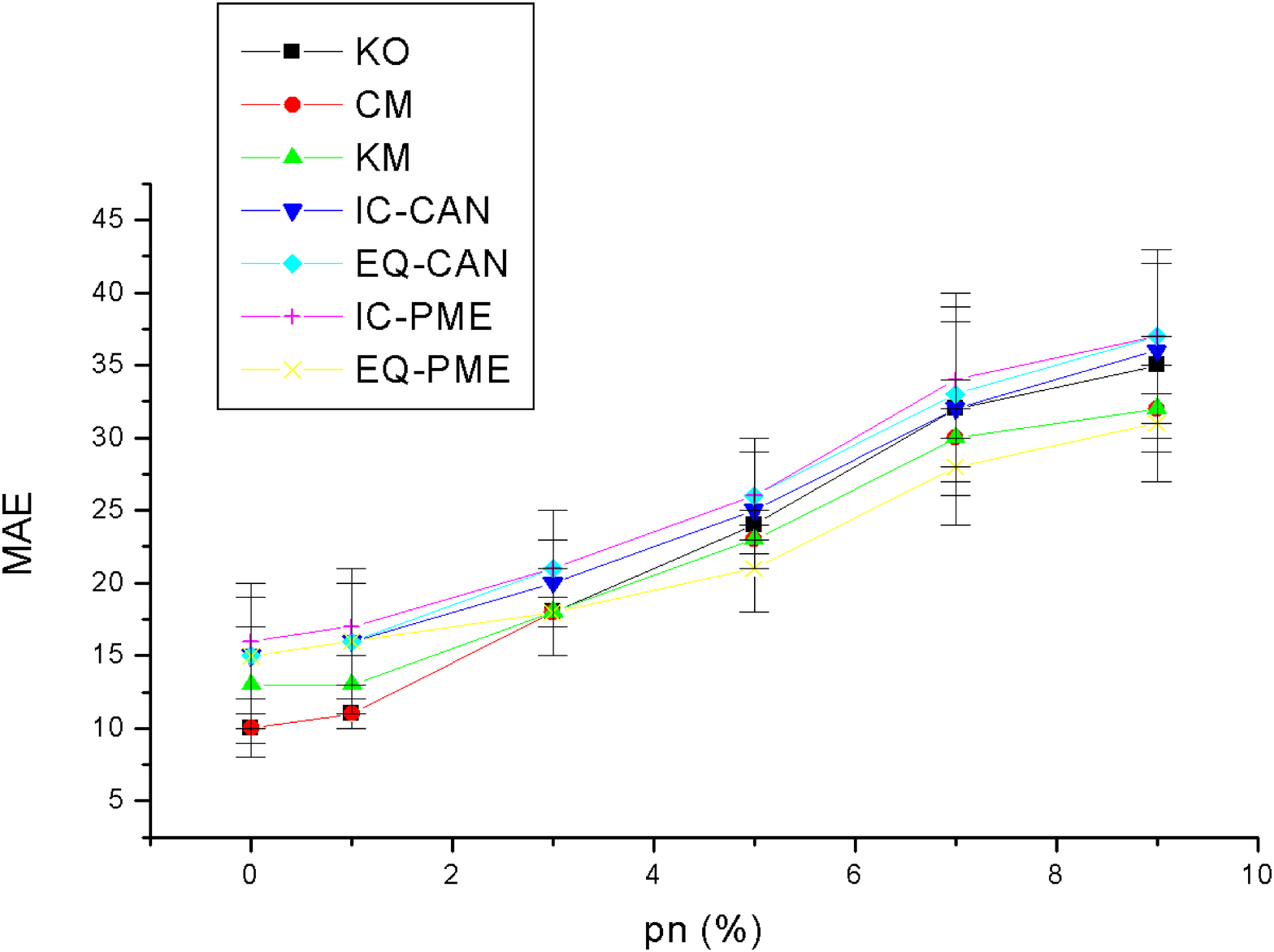}
	\caption{Resultados de $\epsilon_\textnormal{MAE}$ em função do ruído percentual (ODM)}
	\label{fig:MAEotm}
\end{figure}

\begin{figure}
	\centering
		\includegraphics[width=0.7\textwidth]{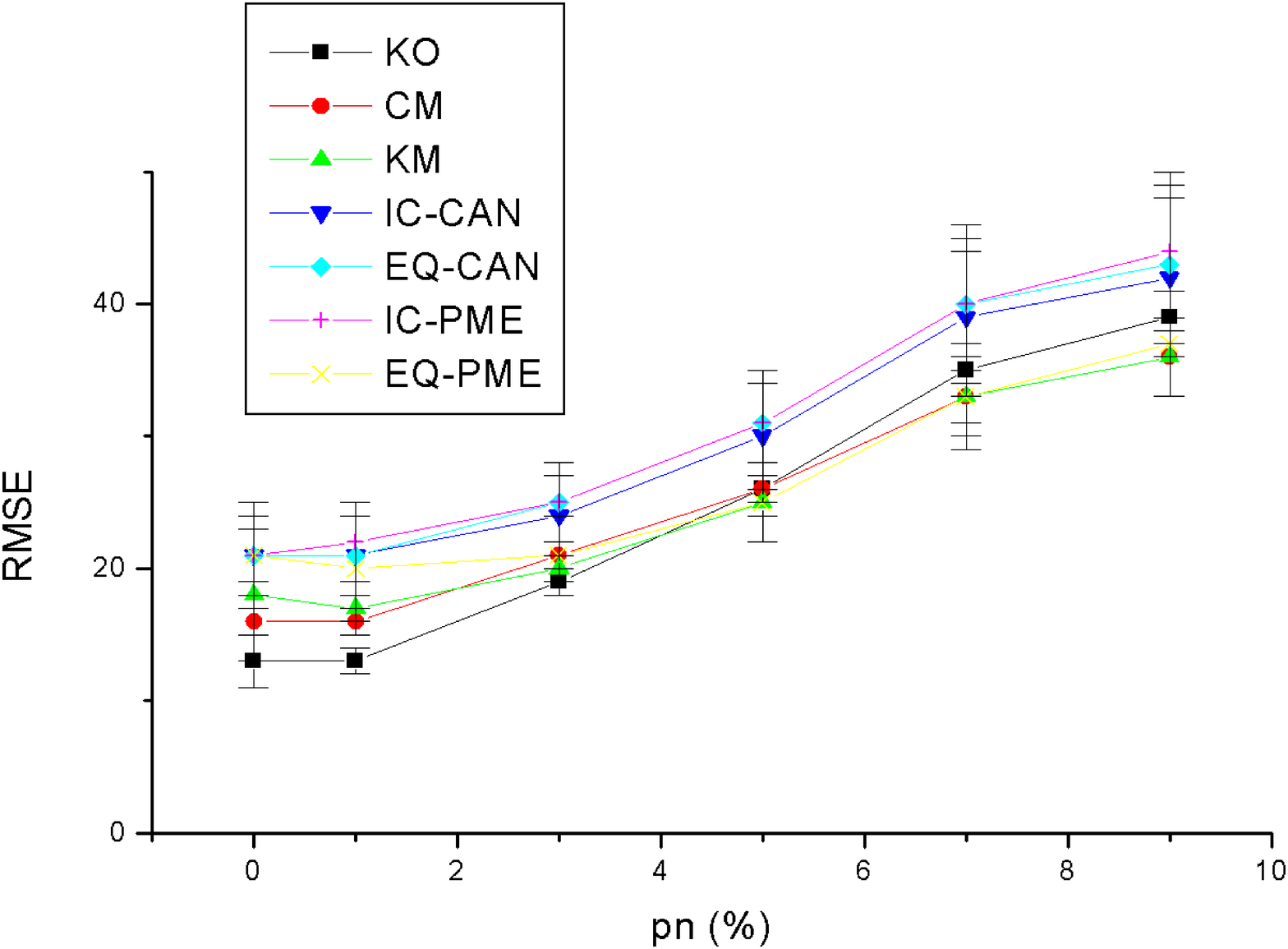}
	\caption{Resultados de $\epsilon_\textnormal{RMSE}$ em função do ruído percentual (ODM)}
	\label{fig:RMSEotm}
\end{figure}

\begin{figure}
	\centering
		\includegraphics[width=0.7\textwidth]{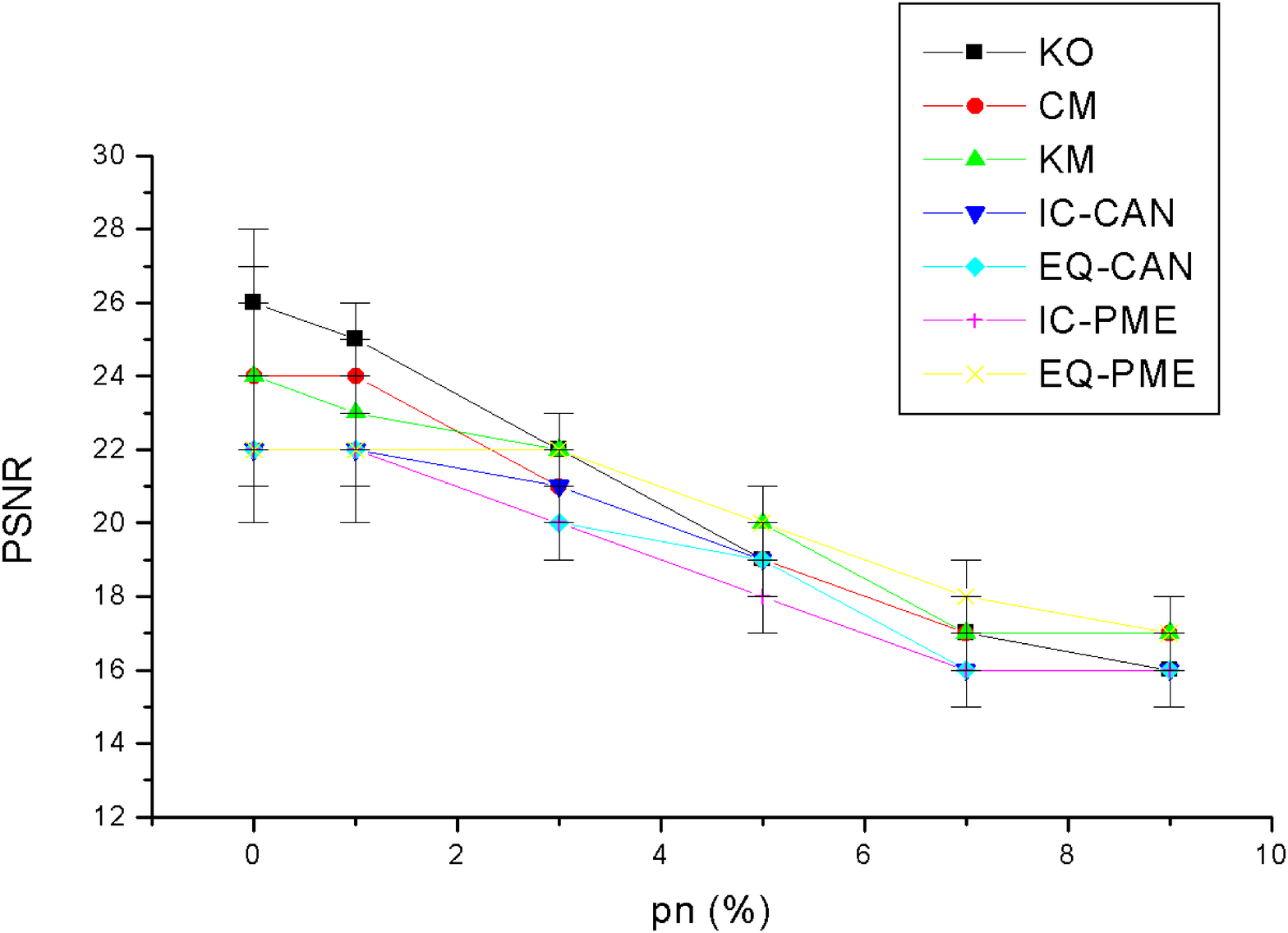}
	\caption{Resultados de $\epsilon_\textnormal{PSNR}$ em função do ruído percentual (ODM)}
	\label{fig:PSNRotm}
\end{figure}

\begin{figure}
	\centering
		\includegraphics[width=0.7\textwidth]{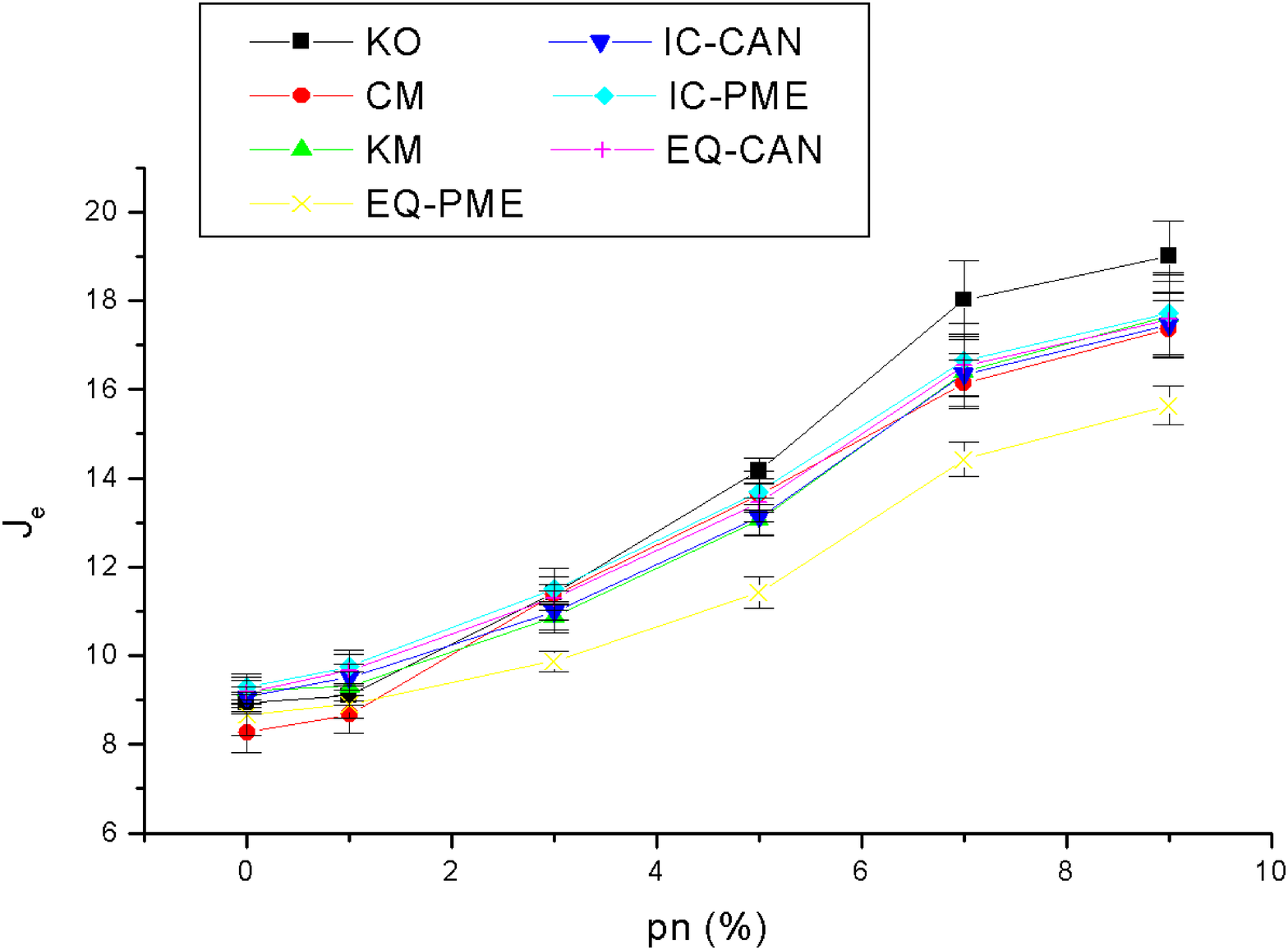}
	\caption{Resultados de $J_e$ em função do ruído percentual}
	\label{fig:EQotm}
\end{figure}

\begin{figure}
	\centering
		\includegraphics[width=0.7\textwidth]{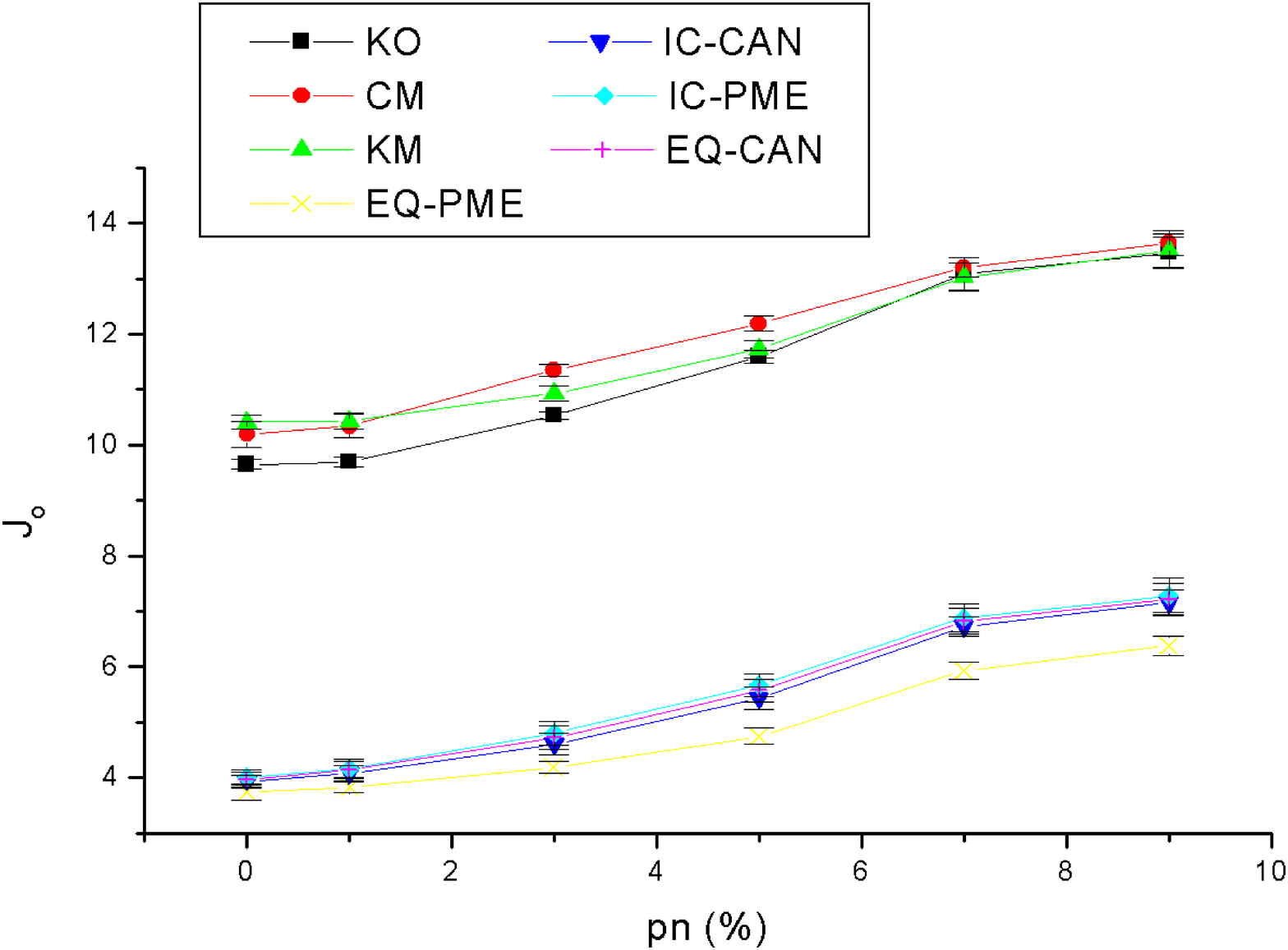}
	\caption{Resultados de $J_o$ em função do ruído percentual}
	\label{fig:ICotm}
\end{figure}

\begin{figure}
  \centering
  \begin{minipage}[b]{0.48\linewidth}
    \centering
    \includegraphics[width=0.5\linewidth]{97_normal_pn0_rf0.eps}
    \\(a)\\
	  \includegraphics[width=0.5\linewidth]{97_class_normal_pn0_rf0_KO.eps}
    \\(b)\\
	  \includegraphics[width=0.5\linewidth]{97_class_normal_pn0_rf0_CM.eps}
    \\(c)\\
    \includegraphics[width=0.5\linewidth]{97_class_normal_pn0_rf0_KM.eps}
    \\(d)\\
  \end{minipage}
  \begin{minipage}[b]{0.48\linewidth}
    \centering
	  \includegraphics[width=0.5\linewidth]{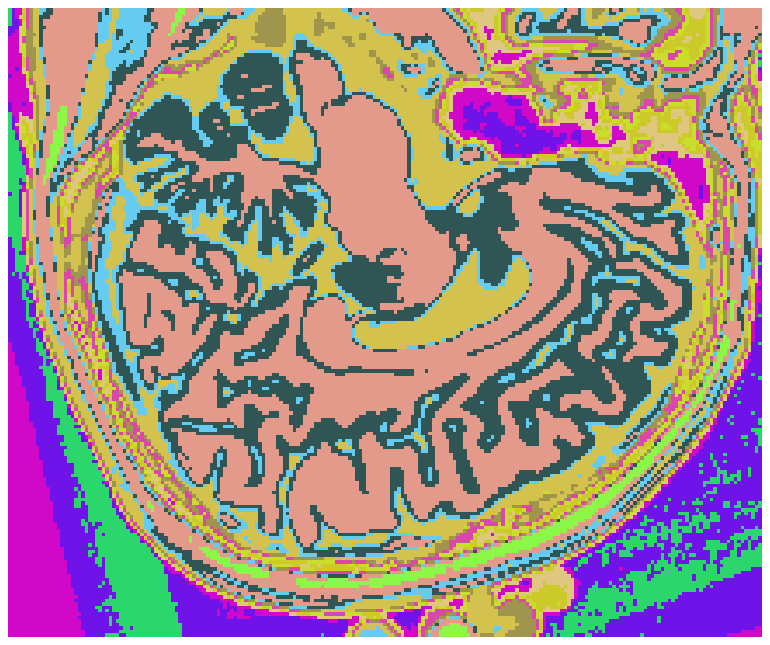}
    \\(e)\\
	  \includegraphics[width=0.5\linewidth]{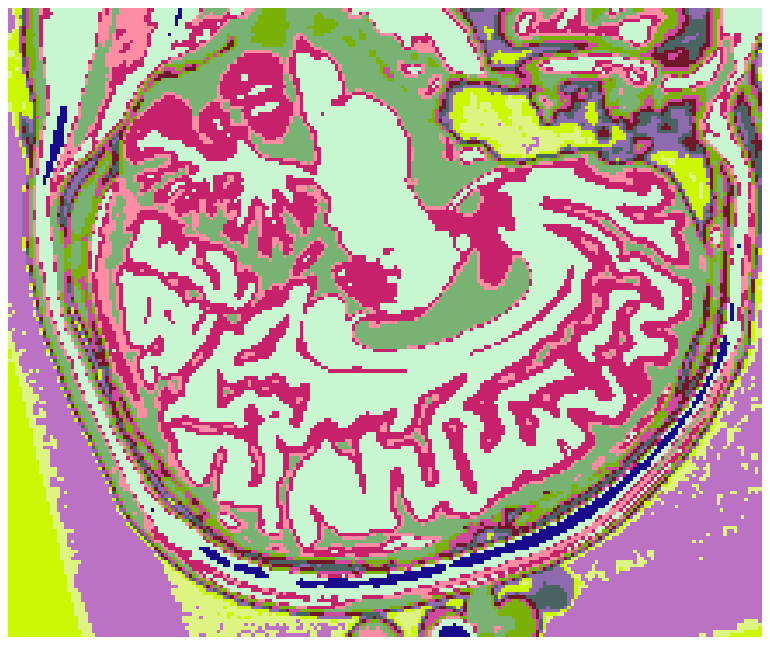}
    \\(f)\\
	  \includegraphics[width=0.5\linewidth]{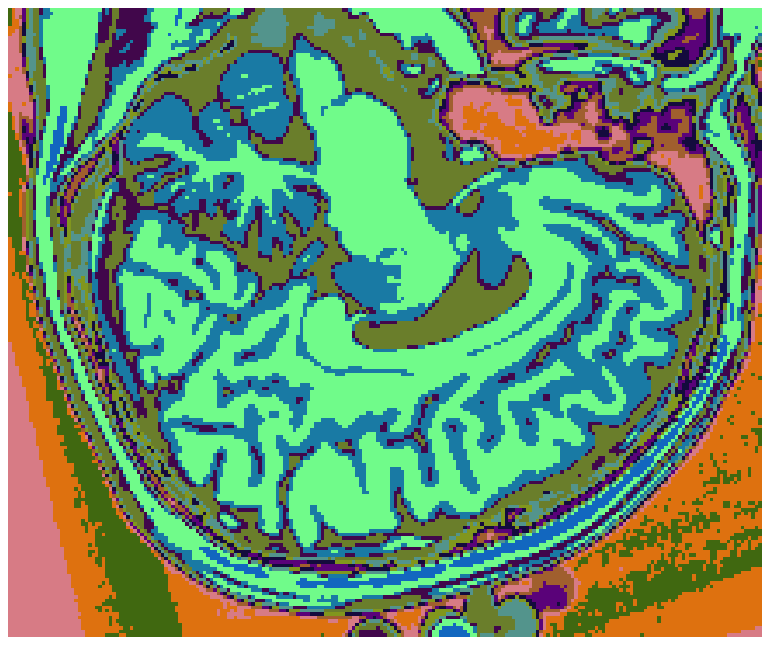}
    \\(g)\\
	  \includegraphics[width=0.5\linewidth]{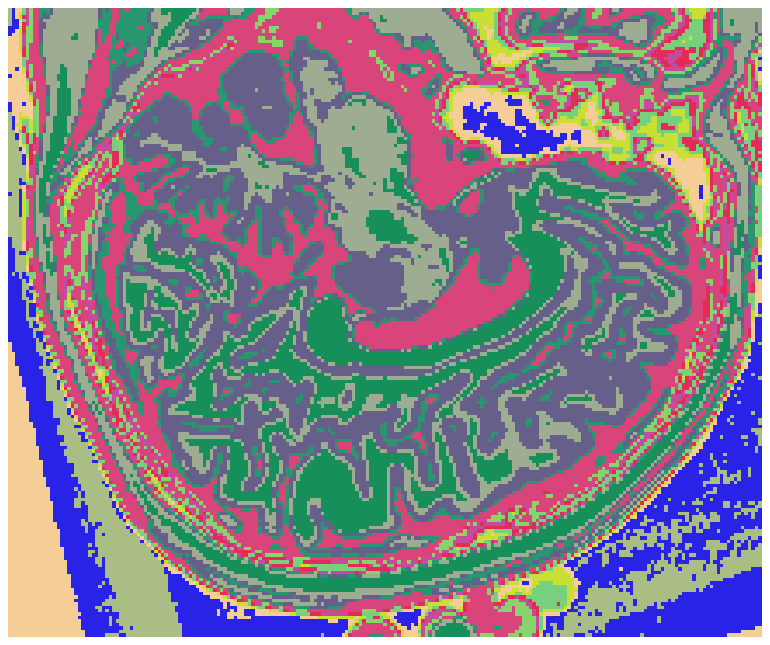}
    \\(h)\\
  \end{minipage}
  \caption{Composição colorida R0-G1-B2 das imagens da fatia 97 ponderadas em PD, $T_1$ e $T_2$ (a) e resultados de classificação usando os métodos KO (b), CM (c), KM (d), IC-CAN-KM (e), IC-PME-KM (f), EQ-CAN-KM (g) e EQ-PME-KM (h)}
  \label{fig:resultados_classificacoesODM}
\end{figure}

\begin{figure}
  \centering
  \begin{minipage}[b]{0.48\linewidth}
    \centering
    \includegraphics[width=0.5\linewidth]{97_normal_pn0_rf0.eps}
    \\(a)\\
	  \includegraphics[width=0.5\linewidth]{97_quant_normal_pn0_rf0_KO.eps}
    \\(b)\\
	  \includegraphics[width=0.5\linewidth]{97_quant_normal_pn0_rf0_CM.eps}
    \\(c)\\
    \includegraphics[width=0.5\linewidth]{97_quant_normal_pn0_rf0_KM.eps}
    \\(d)\\
  \end{minipage}
  \begin{minipage}[b]{0.48\linewidth}
    \centering
	  \includegraphics[width=0.5\linewidth]{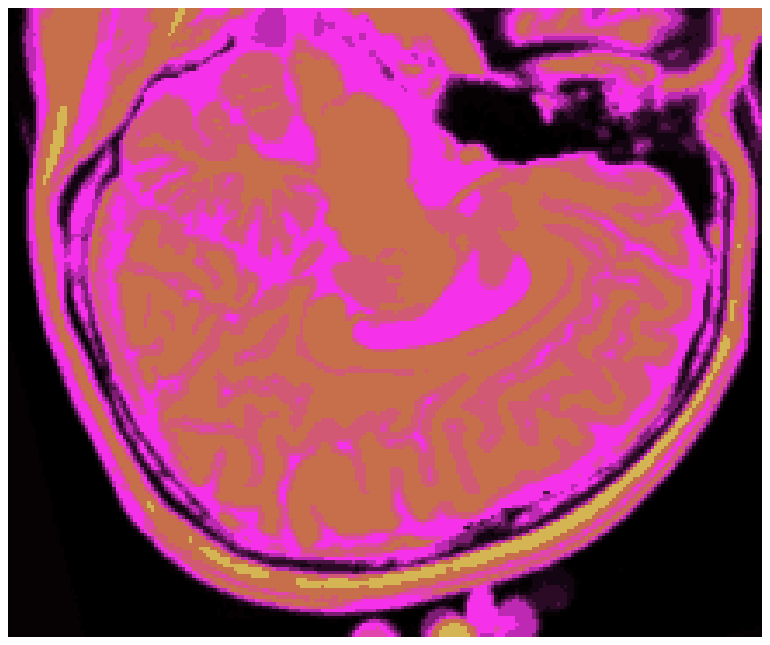}
    \\(e)\\
	  \includegraphics[width=0.5\linewidth]{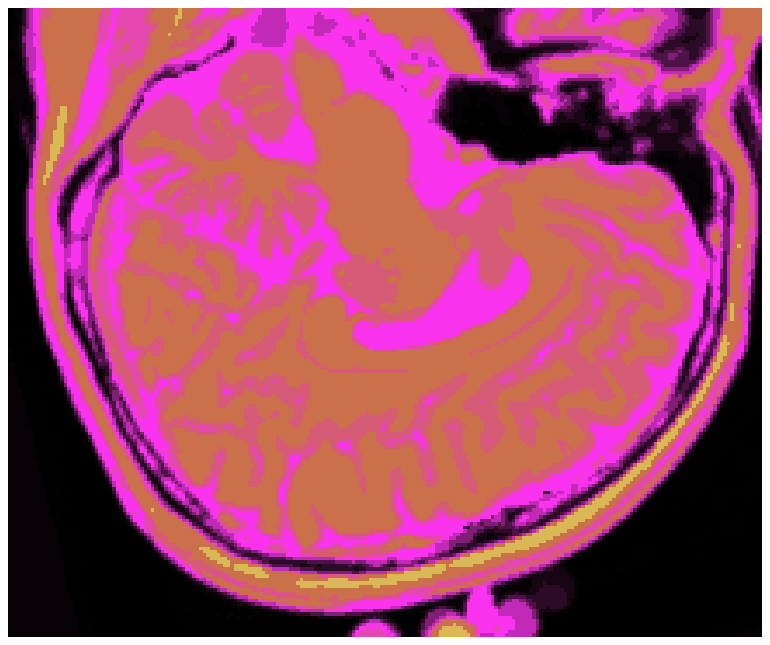}
    \\(f)\\
	  \includegraphics[width=0.5\linewidth]{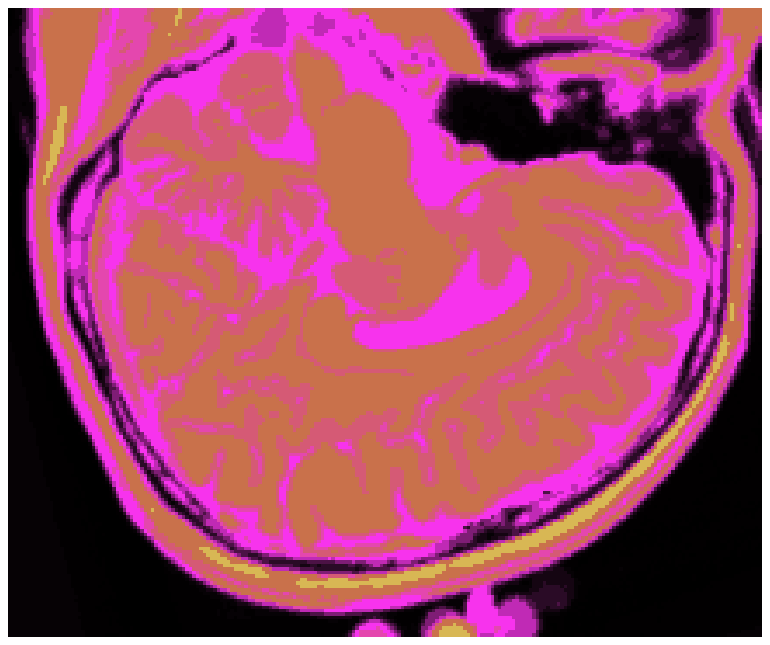}
    \\(g)\\
	  \includegraphics[width=0.5\linewidth]{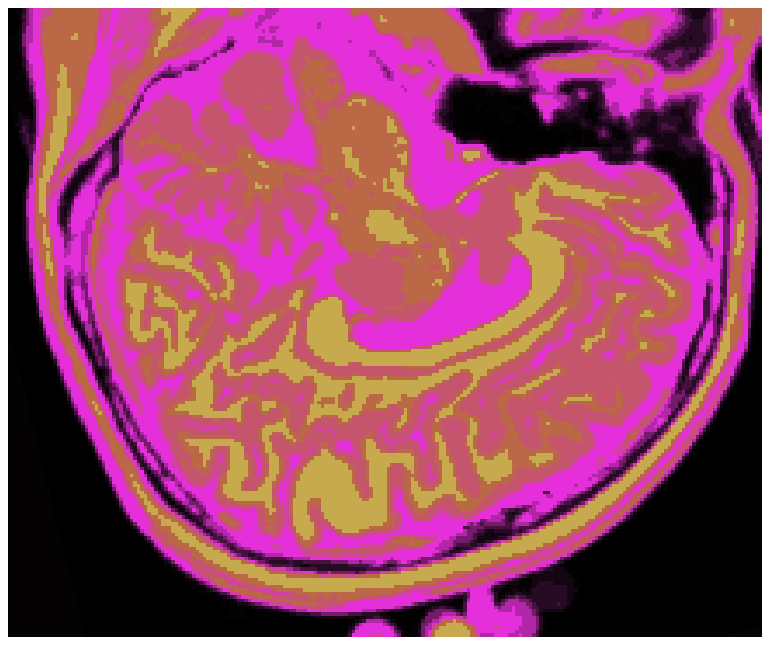}
    \\(h)\\
  \end{minipage}
  \caption{Composição colorida R0-G1-B2 das imagens da fatia 97 ponderadas em PD, $T_1$ e $T_2$ (a) e resultados de quantização usando os métodos KO (b), CM (c), KM (d), IC-CAN-KM (e), IC-PME-KM (f), EQ-CAN-KM (g) e EQ-PME-KM (h)}
  \label{fig:resultados_quantizacoesODM}
\end{figure}

Os gráficos das figuras \ref{fig:MEotm}, \ref{fig:MAEotm}, \ref{fig:RMSEotm} e \ref{fig:PSNRotm} mostram os resultados das medidas dos índices de fidelidade $\epsilon_\textnormal{ME}$, $\epsilon_\textnormal{MAE}$, $\epsilon_\textnormal{RMSE}$ e $\epsilon_\textnormal{PSNR}$, enquanto os gráficos das figuras \ref{fig:EQotm} e \ref{fig:ICotm} exibem os resultados do erro de quantização, $J_e$, e do índice de Omran, $J_o$, para os métodos KO, CM, KM, IC-CAN, EQ-CAN, IC-PME e EQ-PME, em função dos níveis percentuais de ruído, para os 6 volumes de 181 fatias de 3 bandas. A figura \ref{fig:resultados_classificacoesODM} mostra os resultados de classificação, enquanto a figura \ref{fig:resultados_quantizacoesODM} exibe os resultados de quantização para a imagem sem ruído da fatia 97, figura \ref{fig:97_normal_pn0_rf0}, usando os métodos IC-CAN-KM, IC-PME-KM, EQ-CAN-KM e EQ-PME-KM. É possível perceber a partir das imagens que a introdução da otimização dialética permitiu ao k-médias identificar diferentes classes dentro do fundo da imagem.

O gráfico da figura \ref{fig:MEotm} mostra que o método EQ-PME, ou seja, o mapa de k-médias otimizado pelo método dialético de entropia maximizada em função do erro de quantização $J_e$, foi o que apresentou melhor resultado quanto ao índice de fidelidade $\epsilon_\textnormal{ME}$.

Já o gráfico da figura \ref{fig:MAEotm} mostra que os melhores resultados do índice de fidelidade $\epsilon_\textnormal{MAE}$ são aqueles obtidos pelos métodos KO e CM, de 0\% a 3\% de ruído percentual, mas a partir de 3\% o método EQ-PME passa a apresentar os melhores resultados, mostrando que o mapa auto-organizado de Kohonen e o mapa \emph{fuzzy} c-médias são pouco robustos em relação ao ruído, enquanto o mapa de k-médias otimizado pelo método dialético com máxima entropia em função do erro de quantização tem desempenho menos sensível aos níveis de ruído; esse resultado, contudo, não é muito difererente do resultado obtido pelo uso do mapa de k-médias.

O gráfico da figura \ref{fig:RMSEotm} mostra que o mapa auto-organizado de Kohonen apresentou os melhores resultados do índice de fidelidade $\epsilon_\textnormal{RMSE}$ para níveis de ruído percentual de 0\% a 3\%. Contudo, a partir de 3\% os resultados obtidos pelo método KO simplesmente não são mais distinguíveis dos resultados para os outros métodos. Já gráfico da figura \ref{fig:PSNRotm} ilustra que os resultados da aplicação de todos os métodos não são estatisticamente distintos entre si de acordo com o índice de fidelidade $\epsilon_\textnormal{RMSE}$, e isso levando em conta todos os níveis de ruído percentual, mas há uma leve indicação de superioridade dos resultados obtidos com o método de k-médias otimizado usando o método dialético com entropia otimizada em função do erro de quantização, EQ-PME.

Entretanto, o gráfico da figura \ref{fig:EQotm} ilustra que, embora os resultados de 0\% a 1\% de ruído mostrem que os resultados não são distinguíveis entre si quanto ao erro de quantização, a partir de 1\% de ruído os valores do erro de quantização para o método EQ-PME são bem menores do que as medidas de erro de quantização para os outros métodos, mostrando que a otimização do mapa de k-médias usando o método dialético com entropia maximizada em função do erro de quantização tem resultados qualitativamente superiores àqueles obtidos com os outros métodos, conforme também a figura \ref{fig:resultados_quantizacoesODM}.

As diferenças entre os métodos KO, CM e KM e os métodos EQ-CAN, EQ-PME, IC-CAN e IC-PME são mais gritantes quanto a avaliação é feita usando o índice combinado de Omran, como ilustra a figura \ref{fig:EQotm}: todos os métodos baseados na otimização do mapa de k-médias por versões do método dialético retornam medidas de $J_o$ que são pelo menos metade das mesmas medidas obtidas pelos outros métodos.

No entanto, da figura \ref{fig:EQotm} também fica claro que, embora todos os métodos baseados na otimização pelo método dialético retornem medidas de $J_o$ muito melhores que aquelas obtidas usando KO, CM e KM, o método EQ-PME é o que apresenta o melhor resultado dentre todos, destacando-se dos outros métodos a partir de 1\% de ruído percentual, ou seja: o método baseado no mapa de k-médias otimizado usando o método dialético de entropia maximizada em função do erro de quantização $J_e$. Esse resultado não deixa de ser interessante e curioso, pois mostra que o índice de Omran $J_o$, no caso das imagens utilizadas, é dominado pelo erro de quantização $J_e$, dado que a otimização em função do índice de Omran usando os métodos dialéticos canônico (IC-CAN) e com entropia maximizada (IC-PME) são idênticos e qualitativamente inferiores, assim como os resultados obtidos usando o método dialético canônico em função de $J_e$ (EQ-CAN). Isso também indica que a aplicação do Princípio da Máxima Entropia diferenciou sensivelmente o algoritmo de sua versão canônica, tendo acelerado a convergência para a otimização em função do erro de quantização, como bem atestam os resultados.

Esses resultados, combinados àqueles apresentados anteriormente, mostram que, para a aplicação ilustrada neste trabalho, o método EQ-CAN foi superior aos outros métodos de classificação não supervisionada e quantização apresentados, tanto quando são utilizados índices de fidelidade quanto índices de validade do agrupamento, o que mostra que, para as imagens utilizadas, tanto existe um predomínio do erro de quantização $J_e$ sobre o índice combinado de Omran $J_o$ (onde este último, por sua vez, contém o primeiro conforme expressões \ref{eq:omranindex} e \ref{eq:errorquantindex}), quanto existe uma forte influência da aplicação do Princípio da Máxima Entropia ao método dialético de busca e otimização, que nesta aplicação ajudou o método a ter sua convergência acelerada ou até mesmo a vencer um mínimo local.


\section{Conclusões} \label{sec:quanti_conclusoes}

Este trabalho mostrou duas aplicações do método dialético: classificadores dialéticos objetivos e classificadores baseados na otimização dialética. Os classificadores dialéticos objetivos são uma implementação direta do método dialético como definido neste trabalho para problemas de classificação e agrupamento. Foi visto que esses classificadores podem chegar a atingir resultados tão bons quanto os resultados atingidos usando o mapa auto-organizado de Kohonen, considerado um método ótimo para quantização vetorial, com a vantagem de não ser necessário especificar o número exato de classes no início do processo de treinamento, fazendo com que o método seja executado até que se atinja um bom número de classes, necessariamente menor ou igual ao número inicial de classes definidas no início do treinamento.

Já os classificadores baseados na otimização dialética mostraram que é possível obter classificadores otimizados segundo algum critério usando as diversas versões do método dialético e, com eles, conseguir resultados de quantização e de fidelidade superiores àqueles obtidos utilizando o método com parâmetros gerados pela utilização de seu algoritmo próprio de treinamento. Esses classificadores otimizados, quando comparados aos que não fazem uso de otimização e sim utilizam seus próprios algoritmos de treinamento, podem ser uma importante alternativa para evitar com mais facilidade que o processo fique preso em mínimos locais, dada a boa capacidade do método dialético de evitar mínimos locais, como se percebe das situações onde a função objetivo é multimodal, com destaque para a versão com entropia maximizada.

Quanto aos classificadores dialéticos objetivos, os resultados mostram que os classificadores dialéticos e o mapa de kohonen são os que atingem os melhores resultados para 0\% de ruído. No entanto, os resultados obtidos pelos métodos de classificação utilizados não são distinguíveis, não sendo possível comparar os métodos entre si usando índices de fidelidade. Utilizando o teste de $\chi^2$ entre os classificadores dialéticos e os outros métodos, foi mostrado que, para 0\% de ruído, os classificadores dialéticos e o mapa de Kohonen são praticamente idênticos (similaridade acima de 96\%), principalmente para a versão com entropia maximizada. Esses resultados mostram que esses classificadores podem atingir resultados de segmentação e quantização similares àqueles obtidos usando mapas auto-organizados de Kohonen, o que é muito importante, já que os mapas de Kohonen são quantizadores vetoriais ótimos, com a vantagem de os classificadores dialéticos objetivos terem número de classes adaptável, não sendo necessário conhecer o número de classes presente na imagem.

Quanto aos mapas de k-médias otimizados pelos métodos dialéticos de otimização, os resultados mostram que o mapa de k-médias otimizado pelo método dialético de entropia maximizada em função do erro de quantização foi o que apresentou melhores resultados quanto aos índices de fidelidade a partir de 3\% de ruído. A partir de 3\% os resultados obtidos com o mapa de Kohonen não são mais distinguíveis. A partir de 1\% de ruído os valores do erro de quantização para o mapa de k-médias otimizado pelo método dialético de entropia maximizada em função do erro de quantização são bem menores do que as medidas de erro de quantização para os outros métodos, mostrando que tem resultados qualitativamente superiores àqueles obtidos com os outros métodos.

As diferenças entre os métodos são realçadas ao se usar o índice combinado de Omran. Todos os métodos baseados na otimização pelo método dialético resultaram em medidas muito melhores que aquelas obtidas usando os outros métodos. Isso indica que a aplicação do Princípio da Máxima Entropia diferenciou sensivelmente o algoritmo de sua versão canônica, tendo acelerado a convergência para a otimização em função do erro de quantização, como bem atestam os resultados. Entretanto, para a aplicação ilustrada neste trabalho, o método EQ-CAN foi superior aos outros métodos apresentados, tanto quando são utilizados índices de fidelidade quanto índices de validade do agrupamento.

Como trabalho futuro propõe-se a avaliação comparativa do classificador dialético objetivo, em suas versões canônica e otimizada pelo Princípio da Máxima Entropia, e do método de agrupamento de x-médias quanto ao número de classes presente no conjunto de dados de entrada. Como estudos de caso, poderiam ser utilizadas imagens de natureza diversa, com números de classes conhecidos. Outros trabalhos futuros podem envolver o agrupamento de dados numéricos multidimensionais, compostos de classes linearmente e não linearmente separáveis, comparando as diversas versões do classificador dialético com o x-médias, o k-médias, o \emph{fuzzy} c-médias e outros métodos de agrupamento, com ou sem ajuste automático do número de classes.

\renewcommand\refname{Referências}

\bibliographystyle{lnlm}
\bibliography{arq_bib}
\end{document}